\RequirePackage{fixltx2e}
\documentclass[Afour,sageh,times]{sagej}

\usepackage{moreverb,url}
\usepackage{multirow}
\usepackage{makecell}
\usepackage{pdfpages}
\usepackage{subcaption}
\usepackage[colorlinks,bookmarksopen,bookmarksnumbered,citecolor=red,urlcolor=red]{hyperref}
\usepackage[title]{appendix}

\newcommand\BibTeX{{\rmfamily B\kern-.05em \textsc{i\kern-.025em b}\kern-.08em
T\kern-.1667em\lower.7ex\hbox{E}\kern-.125emX}}

\begin{document}

%\runninghead{}

\title{The Unseen Targets of Hate – A Systematic Review of Hateful Communication Datasets}

\author{Zehui Yu\affilnum{1}\affilnum{2}, Indira Sen\affilnum{3}, Dennis Assenmacher\affilnum{1}, Mattia Samory\affilnum{4}, Leon Fr{\"o}hling\affilnum{1}\affilnum{2}, Christina Dahn\affilnum{1}, Debora Nozza\affilnum{5} and Claudia Wagner\affilnum{1}\affilnum{2}\affilnum{6}}

\affiliation{\affilnum{1}GESIS - Leibniz-Institute for the Social Sciences, Germany\\
\affilnum{2}RWTH Aachen, Germany\\
\affilnum{3}University of Konstanz, Germany \\
\affilnum{4}Sapienza University of Rome, Italy \\
\affilnum{5}Bocconi University, Italy\\
\affilnum{6}Complexity Science Hub, Austria}

\corrauth{Zehui Yu}
\email{Zehui.Yu@gesis.org}

\begin{abstract}
\textcolor{red}{Warning: This paper discusses and contains content that is offensive or upsetting.}

Machine learning (ML)-based content moderation tools are essential to keep online spaces free from hateful communication. Yet, ML tools can only be as capable as the quality of the data they are trained on allows them. While there is increasing evidence that they underperform in detecting hateful communications directed towards specific identities and may discriminate against them, we know surprisingly little about the provenance of such bias. To fill this gap, we present a systematic review of the datasets for the automated detection of hateful communication introduced over the past decade, and unpack the quality of the datasets in terms of the identities that they embody: those of the targets of hateful communication that the data curators focused on, as well as those unintentionally included in the datasets. We find, overall, a skewed representation of selected target identities and mismatches between the targets that research conceptualizes and ultimately includes in datasets. Yet, by contextualizing these findings in the language and location of origin of the datasets, we highlight a positive trend towards the broadening and diversification of this research space.

\end{abstract}


\maketitle

\section{Introduction}
Hateful communication is a vehicle of conflict between individuals and groups and exposure to hateful communication online is not a rare phenomenon. In a cross-national survey of internet users, 53\% of American respondents report being exposed to hate material online, while 48\% of Finns, 39\% of Brits, and 31\% of Germans report exposure \citep{Hawdon2017}.  In a more recent study in Germany, about 76\% of respondents said they had been confronted with hateful communication online and 39\% had to deal with online hate very often \citep{forsa2023}. Online platforms increasingly appear to be media of de-civilization since even a low prevalence of hateful content can lead to high exposure rates if uncivil content becomes popular.

Researchers, legal scholars, and practitioners have not agreed upon a single definition of hateful online communication and definitions range from very specific to extremely broad \citep{siegel2020}. Our definition of hateful online communication builds upon the definitions of hate speech by the Encyclopedia of the American Constitution \citep{nockleby2000hate} and Britannica \citep{Curtis2023-ft}. Consequently, we define hateful communication as \begin{quote} 
\textit{``any form of communication or expression (e.g., speech, images, text) that denigrates a person or persons on the basis of (alleged) membership in a social group identified by attributes such as race, ethnicity, gender, sexual orientation, religion, age, physical or mental disability, and others."}
\end{quote}

To tackle the problem of hateful online communication, we need to detect and address it. To this end, practitioners and researchers devoted significant effort to developing automated methods to detect hateful online communication based on Machine Learning (ML).
Since it is well known in Computer Science that the performance of an ML model is upper-bound by the quality of the training data, the topic of data quality gained more attention recently \citep{Abhinav2020, Geiger2021, Liang2022}. 
The famous \textit{``garbage in, garbage out"} principle does not only apply to supervised ML approaches (where the data quality depends on the quality of data annotations that guide the ML model, among other factors) but also to semi-supervised and fully unsupervised ML methods (where the quality of data depends amongst others on the data selection and preprocessing decisions).
Dimensions of data quality that are typically discussed by ML scholars include noisy labels/annotations, class imbalance, data coverage, data homogeneity, and data valuation \citep{Abhinav2020,Liang2022}. In the social sciences scholars differentiate between intrinsic and extrinsic data quality dimensions \citep{daikeler2023assessing}. In the context of hateful communication datasets, intrinsic quality dimensions refer to the extent the dataset covers the phenomenon of interest in its full diversity; extrinsic quality dimensions relate to the accessibility and reusability of datasets. How those identities that shape and are covered by the datasets impact the intrinsic and extrinsic quality of datasets has received little attention so far.

While it is well known that curating datasets requires crucial design decisions that impact their quality, little attention has been paid to the identities of those who curate the datasets. The identity of an individual refers to its community, socio-demographics, position, or self-representation, including but not limited to political affiliation, age, body image, and institutional or organizational membership. Especially when curating data for hateful online communication we expect that the identities of data curators may impact the data quality of the final data in at least two ways: on the one hand, the scientific environment and background of the researchers may affect their definitions of the construct and their practices as data curators; on the other hand, their awareness, interests, and sensibilities towards different targets of hateful communication intersect with their own beliefs, attitudes, and experiences, which in turn may affect their choices on which targets and phenomena to include in the datasets.

Furthermore, issues of data quality related to the identities of researchers compound with those related to annotators. Previous research suggests that identities and beliefs of data annotators impact their perceptions and consequently their annotation of hateful online communication \citep{sap-etal-2022-annotators, pei-jurgens-2023-annotator}. \citet{sap-etal-2022-annotators} find that more conservative annotators and those who scored highly on their scale for racist beliefs were less likely to rate anti-Black language as toxic, but more likely to rate African American English dialect as toxic. More recently \citet{pei-jurgens-2023-annotator} re-annotated 1500 comments sampled from a dataset consisting of Reddit comments \citep{hada-etal-2021-ruddit} using 262 annotators from a representative sample from
prolific. Their results show that people from other
cultures may perceive the same comment with a
lower or higher degree of offensiveness.

For ML models to have a real-world, positive impact on the targets of hateful communication, it is necessary to unpack the relationships between the identities included in the datasets and identities involved in the curation of datasets.

Recent systematic reviews summarized how the literature in hateful communication advanced methodologically and theoretically \citep{Paz2020, vidgen2020directions, PamungkasBP23}. Our research adds to this body of literature by addressing the practices around 
curating hateful communication datasets, with a focus on how they represent the targets of hateful communication. In particular, we offer a positionality outlook on hateful communication research. We perform a systematic review of the past decade of datasets meant for training ML models for detecting hateful language, focusing on the identities that are included and that shape the production of hateful communication datasets. 

First, we focus on the producers of hateful communication datasets and their practices. We find that in the past five years, the field of hateful communication research broadened its geographic borders, became shaped by international collaboration, and increased its coverage of different languages and platforms in the datasets. Yet, the production is still dominated by researchers with U.S.-based affiliations and the majority of datasets are in English.

Next, we focus on the targets of hateful communication that are explicitly included in their design. Leveraging frameworks for assessing the quality of datasets \citep{sen2021total}, we distinguish between explicitly \textit{conceptualized} targets---those who are included in the explicit definition for hateful communication as a construct---and \textit{operationalized} targets---those who are operationalized in the sampling, annotation and/or analysis of the dataset. We find that hand-in-hand with the broadening of the production of hateful communication research, conceptualized and operationalized targets came to include more identities in recent years. However, some target identities, such as age and body image are rarely covered in any of the datasets, which raises concerns about the ability of ML systems to detect hate towards those target identities.

Focusing on a sample of 15 highly-used datasets, we analyze the discrepancy between the targets included in writing---conceptualized and operationalized targets---and the \textit{detected targets}, actually present in the datasets, independently of whether they were included in writing. We find that among the instances for which we detect targets, up to 16\% fall in single target categories that were not conceptualized and/or operationalized first. This may make the hate classifier perform unpredictably on such targets.

In summary, the paper addresses the following questions: 
\begin{description}
\item \textbf{Q1:} In what countries are producers of hateful communication datasets located, what languages are they studying and how are the datasets' qualities evolving?
\item \textbf{Q2:} Which identities are discussed as targets of hateful communication in the scientific literature? 
\item \textbf{Q3:} Which identities are included as targets in hateful communication datasets, even if not explicitly mentioned in the literature? 

\end{description}

Overall, this work highlights a diversification and broadening of the research space around the curation of hateful communication datasets, in terms of both the participants in the scientific field and their attention toward the targets of hateful communication. Within this overall positive trend, the review identifies shortcomings in how research reflects local contexts and identities. Addressing this gap may help the next decade of research in addressing the needs of the targets of hateful communication more accurately and equitably. Thus, this work suggests practical steps for developing standards and practices that ensure the quality of hateful communication datasets.

\section{Related Work}

Creating an ideal dataset for training and evaluating hate speech detection systems is challenging \citep{sodhi2021jibes}. Frequently observed limitations include datasets that are too narrow in their linguistic diversity, with standard English vernaculars being the most studied \citep{ghosh2022sehc}; datasets that are limited to ad-hoc definitions of hateful content, restricting the validity of the resulting machine learning models \citep{hardaker2016real, albadi2018they}; and datasets that skew towards frequently studied targets, disregarding less-frequent but equally consequential ones~\citep{gao2017detecting, del2017hate}. For example, \citet{moy2021hate} analyzed the language discrepancy between English and non-English hate speech datasets and highlighted the importance of non-English datasets for hate speech detection, especially in multi-lingual countries. The study of \citet{swamy2019studying} highlights the redundancy and non-generalisability between datasets for abusive language detection through experiments on cross-dataset training and testing. 

Despite the numerosity of datasets that cover instances of hateful content, only a few studies have focused on their quality. \citet{10.1371/journal.pone.0222194} released a literature review on hate speech research, focusing on research papers from the Web of Science core database published through March 2019. Specifically, they concentrated on mapping broad research indices, prevalent research topics, research hotspots, and significant stakeholders such as organizations and contributing areas. \citet{10.1145/3232676} provided an analysis of the status of hate speech by presenting a summary of approaches, covering algorithms, methodologies, and main features. They also focus on categorizing the different works that aim to detect hate speech for different targets (i.e.,  Racism, Sexism, Prejudice toward refugees, Homophobia, and General hate speech). \citet{vidgen2020directions} reviewed 63 publicly available abusive language datasets also using the PRISMA review methodology. They described the information that the datasets contain (and exclude), how they have been annotated, and how tasks have been constructed. Lastly, they gave a comprehensive examination of methods for making training datasets more accessible and helpful. \citet{DBLP:journals/lre/PolettoBSBP21} systematically assessed the hate speech datasets' characteristics, including their creation methods, thematic focus, and language coverage. While they do not cover targets specifically, they analysed topical focus, i.e., the specific topics and abusive phenomena addressed. For example, topical focuses can be aggressiveness, homophobia, toxicity, or misogyny. While topical focus can also consider targets, this is more related to the task addressed. 

Some reviews on hateful communication paid particular attention to Natural Language Processing (NLP) methods rather than datasets. \citet{schmidt-wiegand-2017-survey} presented a survey on the automatic detection of hate speech, mainly focusing on the NLP approaches. A survey review conducted by \citet{torregrosa2021survey} focused on the existing NLP techniques on extremism detection and their application and mentioned datasets about their availability. Similarly, \citet{jahan2021systematic} systematically reviewed literature of the last 10 years from a technological perspective, with a special focus on NLP and deep learning technologies applied for automatic hate speech detection. \citet{ayo2020machine} focused on the Machine Learning techniques for hate speech classification of Twitter data and provided their current status and future directions.

Unlike previous work, our study focuses on the identities included in the datasets and identities involved in the curation of datasets and how they impact the quality of hateful communication datasets. Target identities play a main role in the dataset from its production to the analysis, which varies a lot across contexts and languages. It could influence the type of content being included and the results of analysis, depending on the research topic of the original study. Prior studies have focused more on explicit characteristics of the dataset being analyzed like language and the approaches used for analysis, while less attention has been paid to the mediator that distinguishes the dataset, namely the target identities of hateful communication. Unlike previous work, our study focuses on the identities included in the datasets and identities involved in the curation of datasets and how they impact the quality of hateful communication datasets.

\begin{figure*}[h]
    \centering
    \includegraphics[width= 0.7\textwidth]{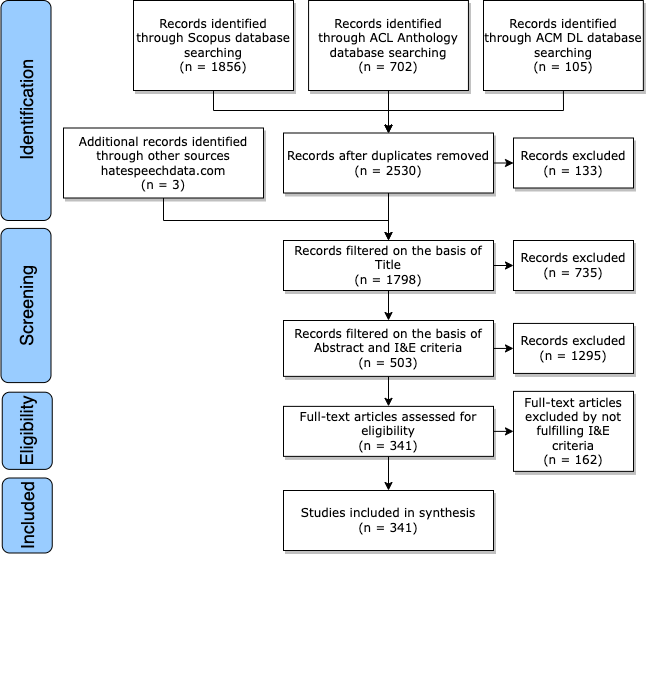}
    \caption{\textbf{Details of literature search and screening process using the PRISMA flow diagram.} Notes: DL stands for Digital Library; I\&E stands for Inclusion and Exclusion. 
    \label{F1}}
\end{figure*}

\begin{figure*}
    \centering
    \includegraphics[width= 0.9\textwidth]{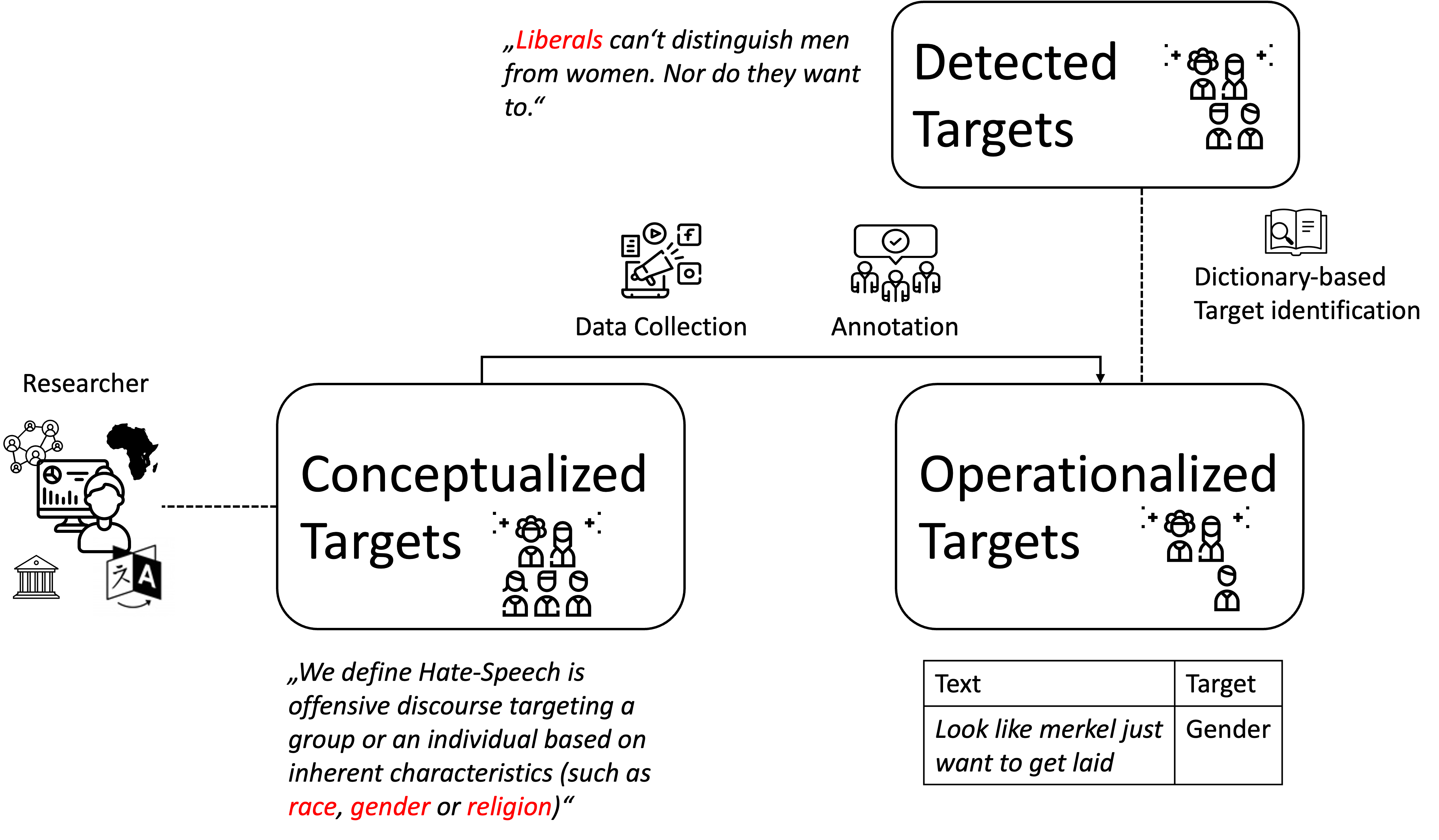}
    \caption{\textbf{The three types of targets studied in this work and the potential mismatches between them.} We introduce a two-tier categorization of targets. First, we distinguish between conceptualized targets (i.e., those who are included in the explicit definition of hateful communication as a construct chosen by the researcher) and operationalized targets (i.e., those who are operationalized in the sampling, annotation and/or analysis of the dataset). Moreover, while conceptualized and operationalized targets are explicitly accounted for and typically described in the paper, the corresponding dataset may include other targets that are not: we call the latter \textit{detected} targets. The figure depicts a mismatch between these three types of targets: the researcher has chosen a very broad conceptualization of hateful online communication encompassing rage, gender, and religion, but a narrow operationalization, which only aims to capture hate towards gender identities in the dataset; yet, ultimately, the final dataset may include also targets that were part neither of the conceptualization nor the operationalization, such as identities based on political ideology.}
    \label{F2}
\end{figure*}

\section{Methods and Data}
\label{sec:methods}

In this section, we outline how we survey the literature and analyze papers that introduce novel datasets of hateful communication. Then, we clarify the methodology used to assess the identities that are included and that shape the hateful communication datasets and their quality.

\subsection{Paper Analysis: Systematic Literature Review}
We follow the PRISMA guidelines for surveying the literature systematically~\citep{page2021prisma}. Here, we clarify how we search, select, and annotate papers. In particular, to gain an encompassing view of the quality of the datasets introduced by this body of work, we annotate the papers introducing the datasets, the datasets themselves, and the targets of hateful communication explicitly mentioned in the papers. We conducted this literature review in early 2022; therefore our sample is restricted to papers published till March 2022. To prevent misunderstandings related to the partial availability of data for the year 2022, we de-emphasize the corresponding results by reducing their opacity in plots.

\subsubsection{Search and Selection Procedures}
We based our literature search on the following academic databases and search engines: \textit{Scopus, ACM Digital Library}, and \textit{ACL Anthology} because of their topical relevance and interdisciplinary nature.

To achieve a broad inclusion of datasets concerned with different dimensions of hateful content, we first constructed a set of queries, composed of four parts. These queries systematically combine four different dimensions: topic, content, dataset, and data source. Our broad understanding of hateful content is reflected particularly in the wide range of keywords in the topic dimension. For each dimension, we defined a set of relevant keywords: topics (e.g., \textit{``hate", ``troll", ``dehumanize"}), content (e.g., \textit{``message", ``speech", ``language"}), datasets (e.g., \textit{``corpora", ``dataset", ``corpus"}) and data source (e.g., \textit{``web", ``internet", ``online"}). While these queries help to include a large number of datasets on hateful content, they also lead to the inclusion of related, but not relevant constructs and datasets. These publications are screened out in the second stage of our extensive manual paper selection procedure. An overview of the literature screening process is given in Figure~\ref{F1}. Further details on the search and selection procedures can be found in the appendix.

\subsubsection{Annotation Procedure}
We constructed a concept matrix (Appendix Figure \ref{F14}) that included details about the publications, corresponding datasets, and targets to assist in guiding our data annotation process. For the final round of reviews, three annotators evaluated the full text of each paper and completed the annotation matrix for all papers included in our sample.

\subsubsection{Annotation: Paper Metadata}
The paper section provides the publication's general meta-data, including its title, journal, country of author affiliation, summary, citation number, publication date, and accessibility. Within the dataset section, we differentiate between construct definition, metadata about the dataset (e.g., its collection procedure, time span, topical focus) and its annotation (e.g., number of annotators, guidelines, incentives).

\subsubsection{Annotation: Dataset Metadata} 
The dataset metadata block contains detailed information about the dataset, including its availability, way to access, format, reference name, number of newly created or adapted datasets, language, data source, topical focus, relevance to social events, data production and collection time, country of origin, and any measures taken for anonymity protection of the data source. We explored the annotation process that was used during the dataset creation by identifying information about the annotation type, procedure, selection strategy, overall data size and the size used for training and testing, information about annotators, guidelines, and incentives provided.

\subsubsection{Annotation: Conceptualized Targets}
The target section provides information about the authors' definitions of targets of hateful communication that are explicitly addressed by the authors of the datasets. We differentiate between papers that aim to measure hate towards selected targets and papers that do not discuss specific targets. We further differentiate between individual targets (e.g., a specific politician), and targets corresponding to collective identities (e.g., one's political affiliation). We categorize collective identity targets according to the taxonomy introduced in \cite{elsherief2018hate} and iteratively added four other major collective identities we found in the publications in our literature review, namely---political affiliation, age, body image, and institutional or organizational membership.\footnote{~\cite{elsherief2018hate} also include an \textit{``archaic"} target category which we did not find in any other publication, hence we dropped it.}

For the construct, we collected its definitions in the text of the papers, including all potential sub-categories of the main construct. From the definitions, we extract both the topical focus---the communicative phenomenon under study, such as male chauvinism---and the targets of the hateful communication---such as women. In this paper, we call \textit{conceptualized targets} those that are explicitly mentioned in the construct definition.

As an example,~\citet{taradhita2021hate} define the hate speech construct as \textit{``an act of communication by a particular person or group that aims to insult a person or a group based on their ethnicity, race, religion, gender, sexual orientation, or class"}, the latter being the targets of the particular kind of hateful communication. On the contrary, some definitions in the literature do not explicitly identify specific groups of targets as part of the construct definition. For example, according to \cite{zhang2018detecting}  \textit{``we identify that hate speech 1) targets individual or groups on the basis of their characteristics (targeting characteristics); 2) demonstrates a clear intention to incite harm, or to promote hatred; 3) may or may not use offensive or profane words"}.

Following frameworks for data quality \citep{sen2021total}, we aim to assess the inclusion of targets in all phases of dataset creation. To this end, we annotate targets beyond those mentioned in the construct definition at two crucial steps: those included intentionally and explicitly in the operationalization (i.e., data creation process), and those included in the final dataset itself even if unintentionally. We clarify the former, before discussing the latter in the next section.

\subsubsection{Annotation: Operationalized Targets}

We annotate as \textit{operationalized targets} those targets for which the authors define concrete measures to ensure that their presence in the data is visible. For example, authors may mention targets in the annotation codebook and/or use them as labels or may use certain methods that are specifically designed to detect certain targets (e.g., a dictionary to detect gender words or an antisemitism classifier). 
Additionally, some authors also define measures that increase the presence of certain targets in the data (i.e., define data collection strategies that potentially boost the presence of certain targets).

\subsection{Dataset Analysis: Detected Targets}
We next turn to the targets that are included in the datasets themselves, irrespective of whether the authors of the datasets mentioned them in the conceptualization or operationalization of the construct under study.  We call these \textit{detected targets}.

Therefore, we complement our literature review with an in-depth analysis of a convenience sub-sample of targeted datasets---a combination of datasets already included in our systematic review and datasets added and annotated exclusively for this analysis---which was made available by \citet{risch2021toxic}. For this sub-sample, we analyse the prevalence of different targets. This analysis allows quantifying the potential mismatch between conceptualized, operationalized and included targets in datasets that are or can be used to train hate speech detection systems. A potential mismatch on these different levels that are depicted in Figure \ref{F2} may lead to surprising failures of hate speech detection systems.

For target detection, we use a dictionary-based approach, which aims to detect different identity terms that are mentioned in a hateful or non-hateful context.

\subsubsection{Dictionary Creation}

Our methodology for the dictionary-based target detection builds on related work with a strong focus on target identities \citep {silva2016analyzing, elsherief2018peer,vidgen2020directions}. Starting with a list of more than 750 keywords from the website hatebase.org, \citet {elsherief2018peer} identify the 51 terms most indicative of hate speech, removing phrases that were deemed context-sensitive or that would frequently be used in contexts other than hate speech, for example, the term \textit{``pancake"}. \url{hatebase.org}, one of the biggest repositories of multilingual hate speech, compiled this list by asking users to contribute through the addition of new hate speech terminology and classifying it into different categories. Both the compressed list of keywords and the categorization scheme were subsequently adopted as a basis for the analysis by \citet{elsherief2018peer}.

After reviewing several works focusing on hate speech targets \citep{davidson2017automated, degibert2018hate, qian2019benchmark, kennedy2020constructing, pamungkas2020you, vidgen2021introducing, kennedy2022introducing}, we decided to further extend the target dictionary proposed by \citet{elsherief2018peer} to improve the coverage of target categories discussed in existing literature. Specifically, we incorporated the categories \textit{Age} and \textit{Body}, along with corresponding keywords found in \citet{vishwamitra2020analyzing} and \citet{baheti2021just}. Additionally, we included the category \textit{Political}, as well as the category \textit{Organizations/Institutions} and corresponding keywords from \citet{zampieri2019predicting}. We excluded the category \textit{Archaic} as it did not represent a single, homogeneous target group.
The final set of target categories is \textit{Age, Body, Class, Disability, Gender, Nationality, Organisations/Institutions (Org./Inst.), Political, Race, Religion}, and \textit{Sexuality}. We report in Table \ref{tab:dictionary} of the appendix the full list of keywords associated with each category.

\subsubsection{Dictionary Application}
For each instance of a dataset, we check whether one or multiple keywords defined in the dictionary occur in the instance's text. To improve the accuracy of these matches and increase the efficiency of the matching process, minor preprocessing steps such as the deletion of stopwords, digits, and punctuation and a transformation into lowercase text were implemented. As the keywords in our dictionary included bigrams, we tokenized the instances' texts into both uni- and bigrams, before applying the dictionaries to them. The results were aggregated on the dataset level, resulting in a distribution of the instances over the available target categories for each dataset. 
To identify the hateful instances in the datasets, we rely on the annotations provided by the dataset creators.

\subsubsection{Dictionary Validation}
While we cannot assess the recall of our dictionary approach for detecting targets, we find that it does afford precision: estimating on a stratified random sample of 33 instances where at least one target was detected (3 instances for each of the 11 target categories), the correct target is present in 21 instances, corresponding to a macro-average precision of 68\%. While the agreement between our annotations and the dictionary results is high for the frequently operationalized target categories, it seems to be inherently difficult to operationalize the categories \textit{Age} and \textit{Body} using a theory-based lexicon. Even though we observe low agreement for these categories, we decided to retain them to surface this difficulty and highlight the lack of datasets covering them. For the categories \textit{Organisations/Institutions} and \textit{Nationality}, the round of manual annotations showed their proximity to the categories \textit{Political} and \textit{Race}, respectively. This observation is in line with the difficulties that \citet{bretschneider2017detecting} and \citet{ousidhoum2019multilingual} report in distinguishing those category pairs during their annotations.

\begin{figure}
    \centering
    \includegraphics[width=.45\textwidth]{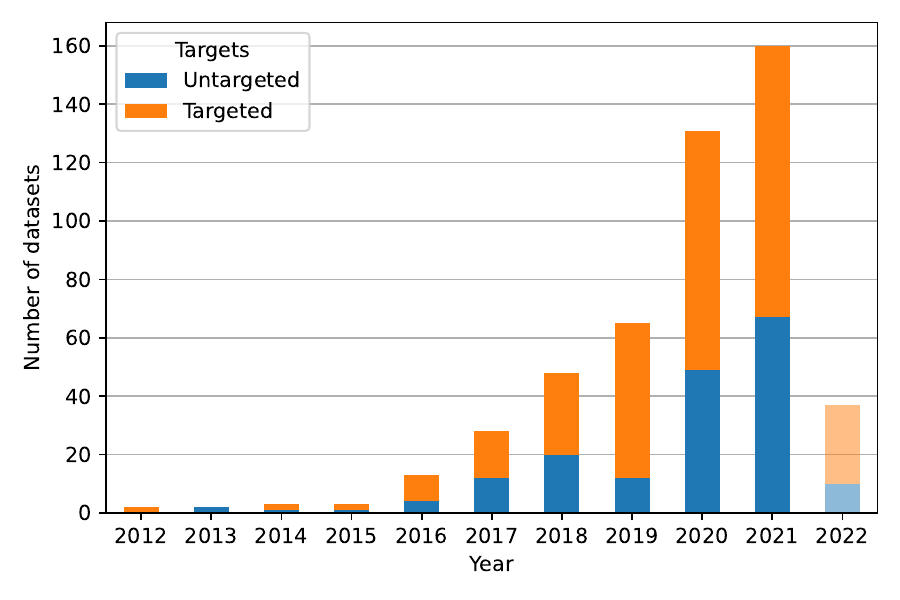}
    \caption{\textbf{Conceptualized and operationalized targets by year along with the distribution of datasets.} Targeted refers to datasets that have explicitly mentioned at least one target in their construct definition (i.e., in the conceptualization phase) and/or publications in which the authors define concrete measures to ensure and validate the presence of at least one target group in the data (i.e., in the operationalization phase). Untargeted refers to all other datasets that do not meet these two criteria. Note that the data for 2022 is only partially available (as described in our Methods and Data Section). }
   \label{F3}
\end{figure}

\begin{figure}
    \centering
    \includegraphics[width=.45\textwidth]{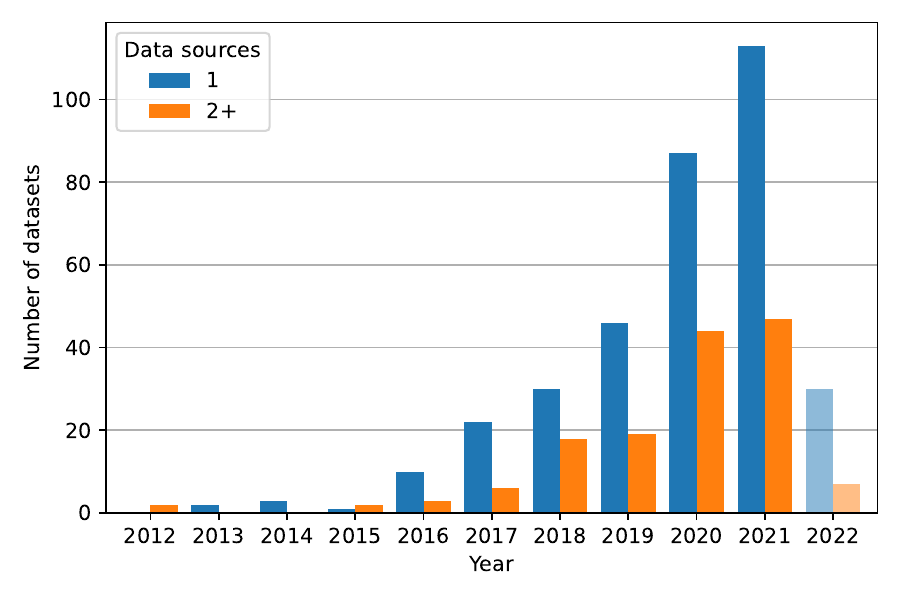}
    \caption{\textbf{Single vs. multiple platforms as data sources over time.} While most of the datasets are collected from a single source, around 2018 researchers are increasingly collecting data from multiple sources (i.e., two or more).}
    \label{F4}
\end{figure}

\begin{figure}
    \centering
    \includegraphics[width=.45\textwidth]{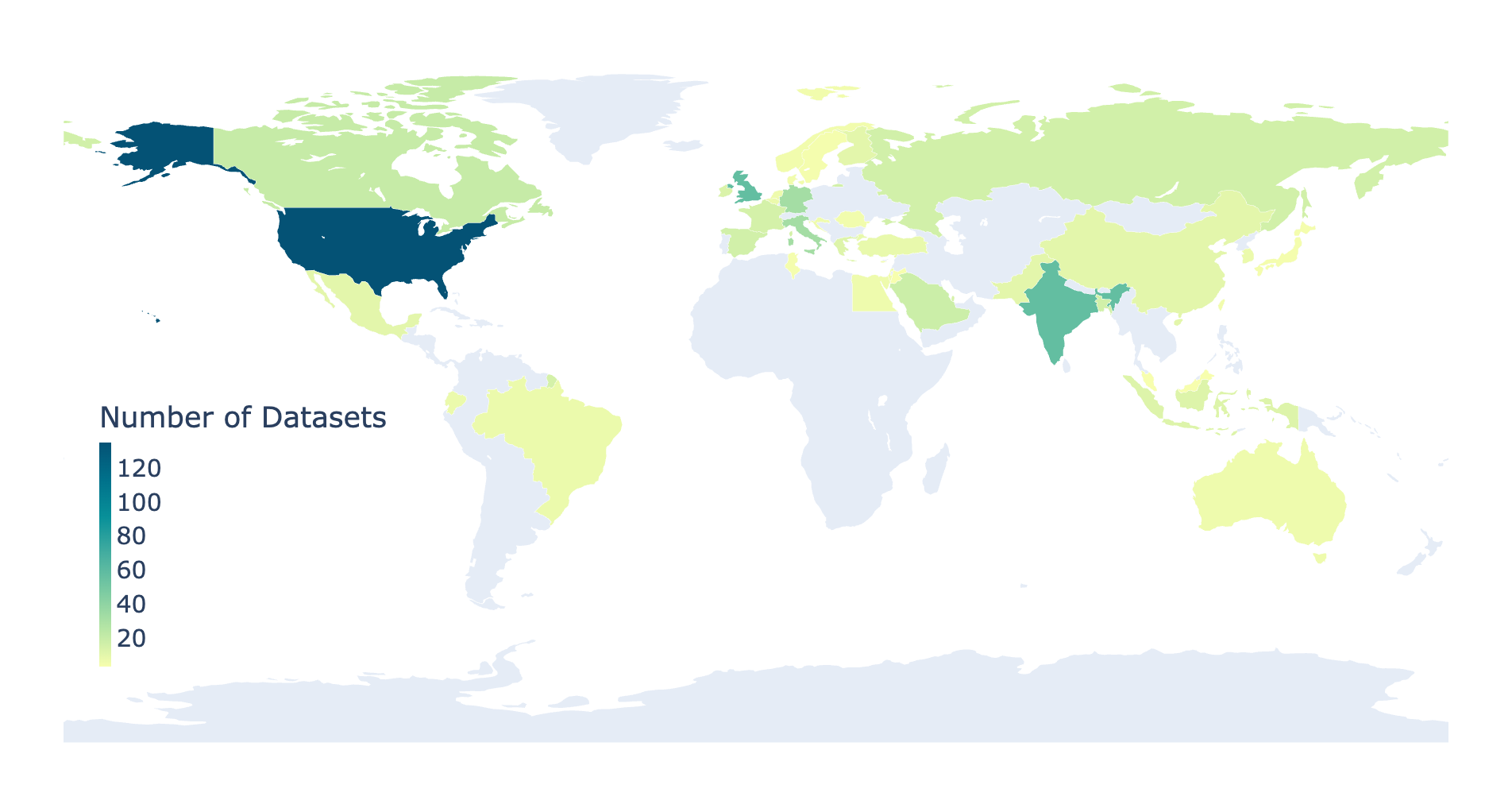}
    \caption{\textbf{Geographic distribution of researchers' affiliation that contributed datasets.} Researchers affiliated with institutions located in the U.S. published the most datasets, followed by researchers from India and the United Kingdom institutions.}
    \label{F5}
\end{figure}

\begin{figure*}
  \centering
  \begin{minipage}[b]{0.45\textwidth}
    \includegraphics[width=\textwidth]{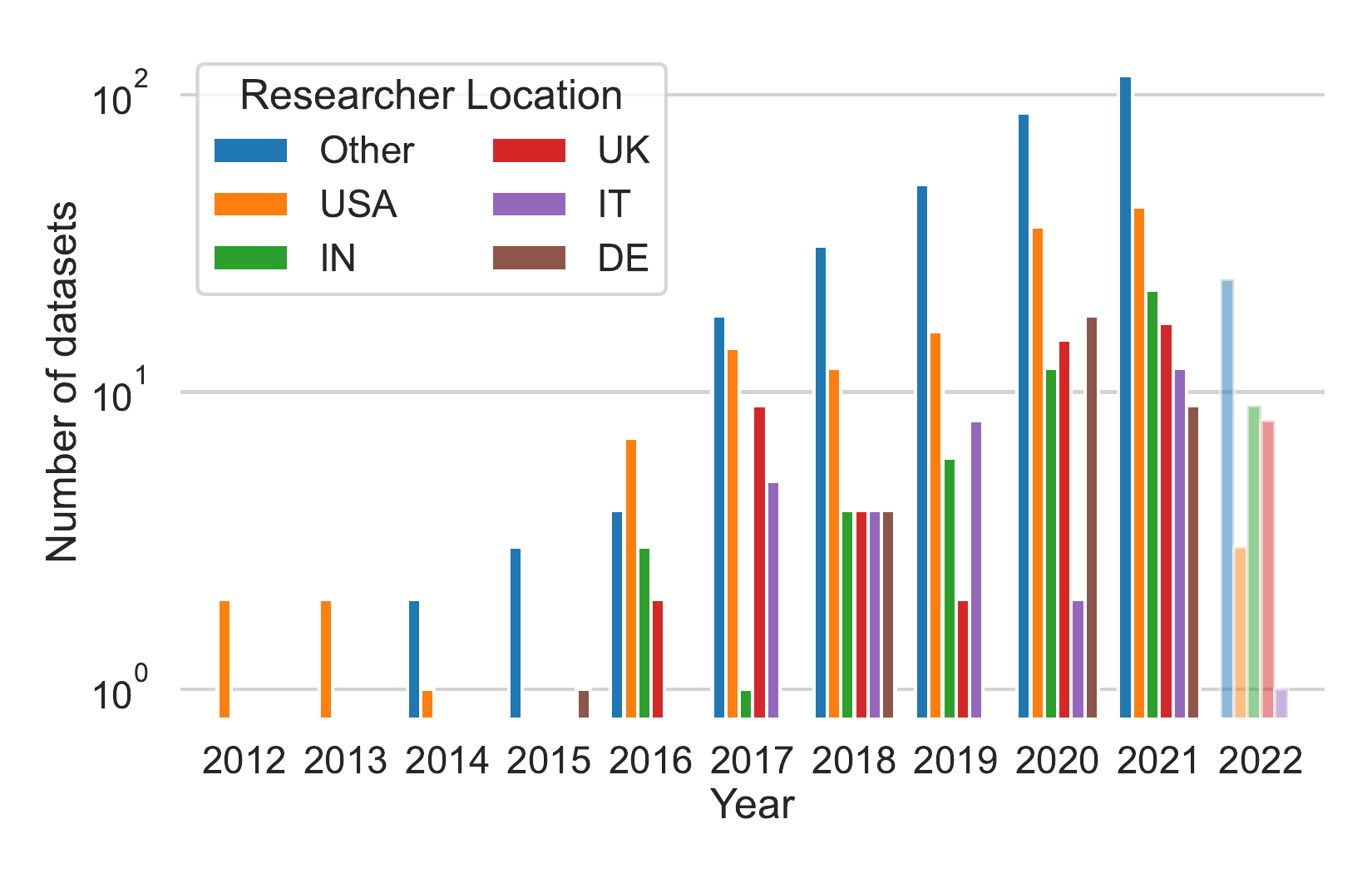}
     \captionof*{figure}{\textbf{(a) The distribution of datasets over time for the five most frequent researcher locations and all other locations combined.} Researchers from the US publish the most consistently, while other countries began producing more datasets since 2016.}
  \end{minipage}
  \hfill % optional: add some horizontal spacing
  \begin{minipage}[b]{0.45\textwidth}
    \includegraphics[width=\textwidth]{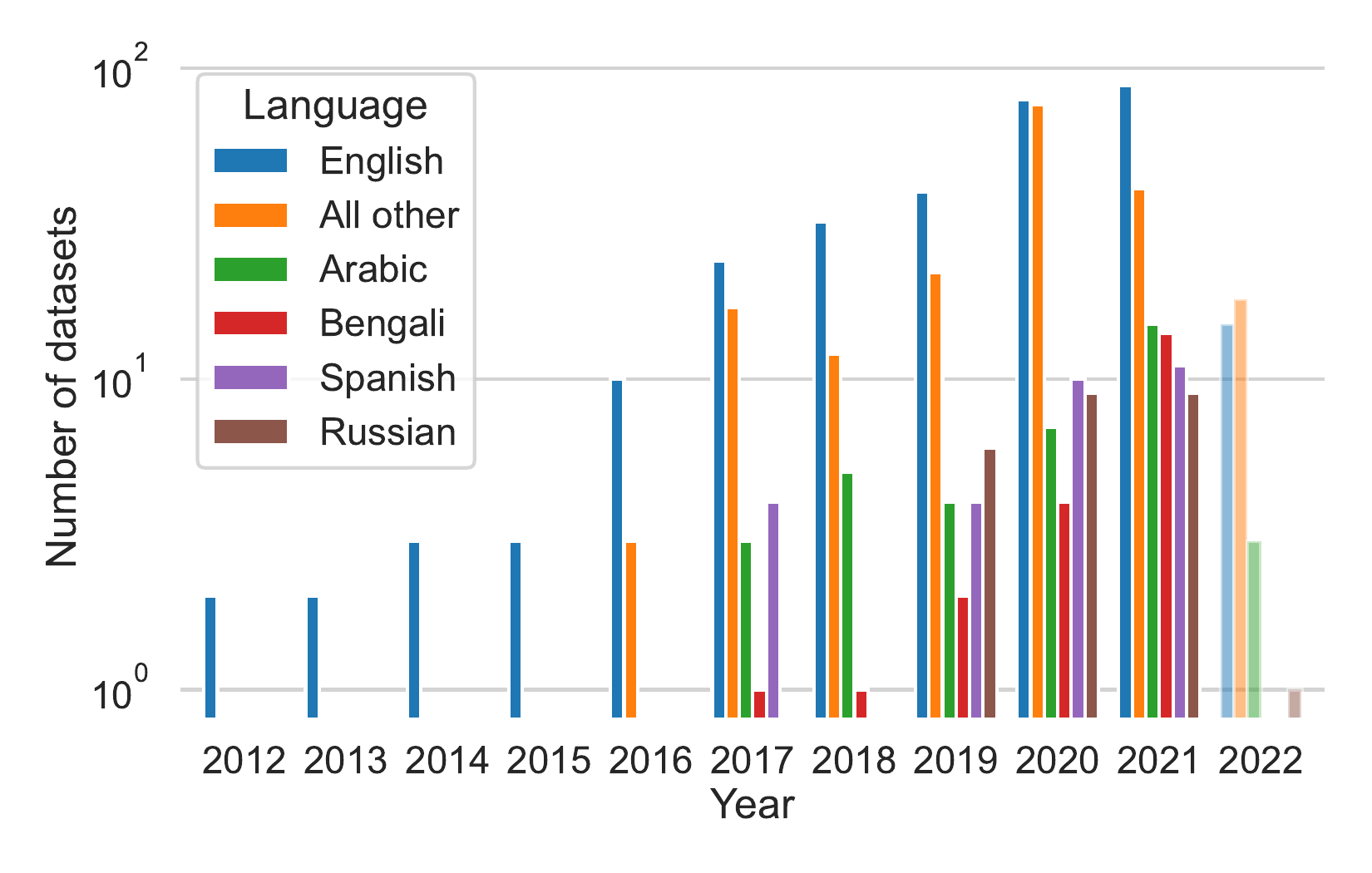}
    \captionof*{figure}{\textbf{(b) The distribution of datasets over time for the five most studied and all other languages combined.} English has received most of the attention which has risen over time, while research on other languages became prominent after 2017.}
  \end{minipage}
     \caption{\textbf{Linguistic and geographic trends over time in harmful language research.} Notably, the number of datasets for the non-top five languages continues to be lower than all English ones, while since 2017 we see more and more datasets in languages other than English.}
    \label{F6}
\end{figure*}

\begin{figure}
    \centering
    \includegraphics[width=.45\textwidth]{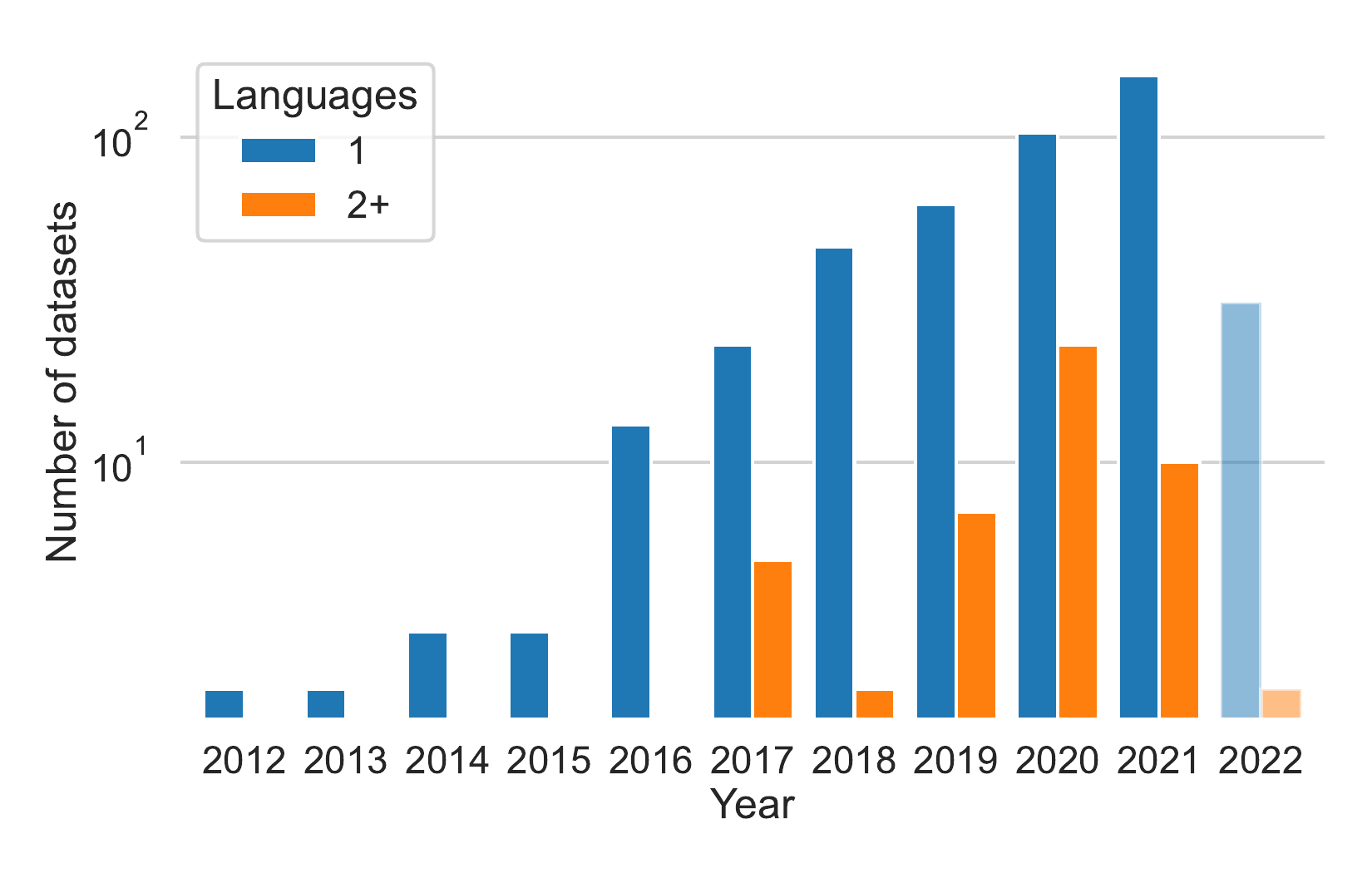}
    \caption{\textbf{Monolingual vs. multilingual datasets over time.} While monolingual datasets are more frequent, around 2017 researchers began producing multilingual datasets.}
    \label{F7}
\end{figure}

\begin{figure*}[h]
    \centering
    \includegraphics[width=\textwidth]{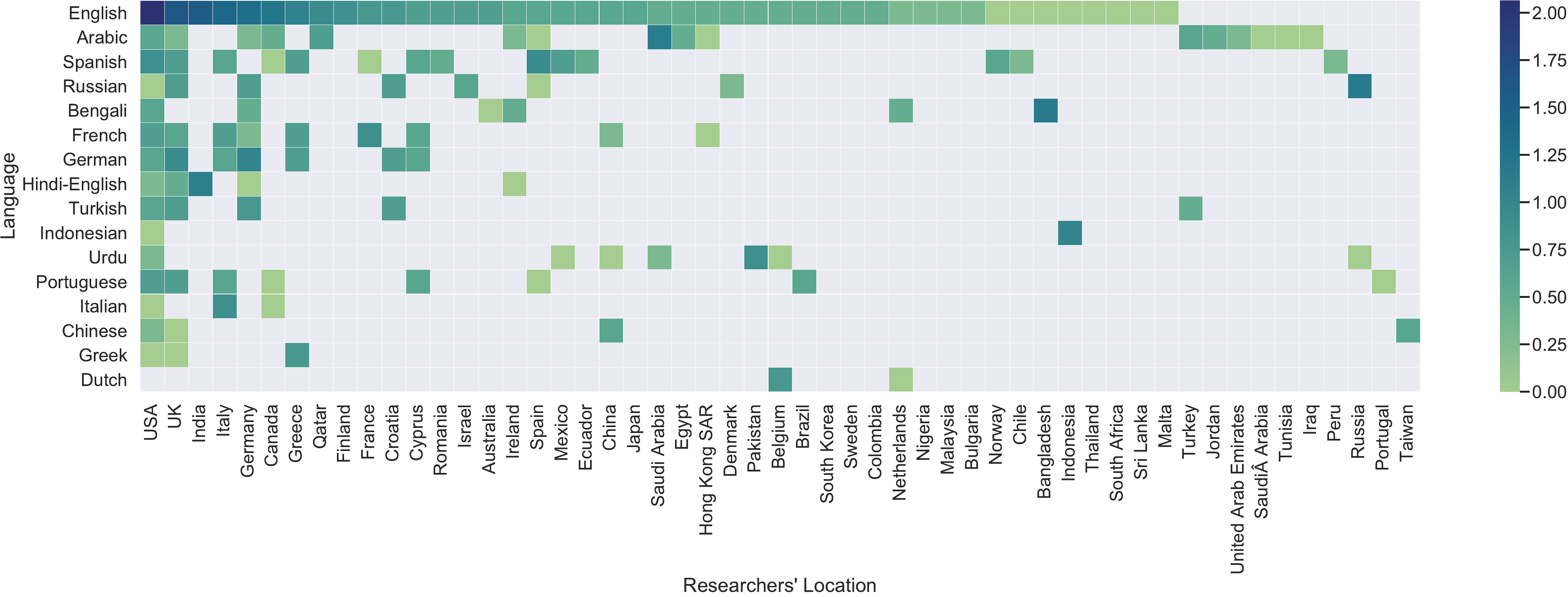}
    \caption{\textbf{Hateful datasets by language and the researchers' location.} We show the distribution of the top 16 language datasets and the 40 most common locations of the researchers who created these datasets. The frequency of datasets is log-scaled to reduce the dominant effect of English language datasets. We see that research in English is widespread across many geographic regions, while researchers from the US and UK contribute to research in a variety of languages. Spanish and Arabic are also researched in multiple countries, reflecting the spread of their worldwide speakers. On the other hand, research in other languages is concentrated in countries or locations where they are most widely spoken, e.g., publications with Indonesian and Russian datasets originate from Indonesia and Russia, respectively.}
    \label{F8}
\end{figure*}

\begin{figure*}[h]
    \centering
    \includegraphics[width=\textwidth]{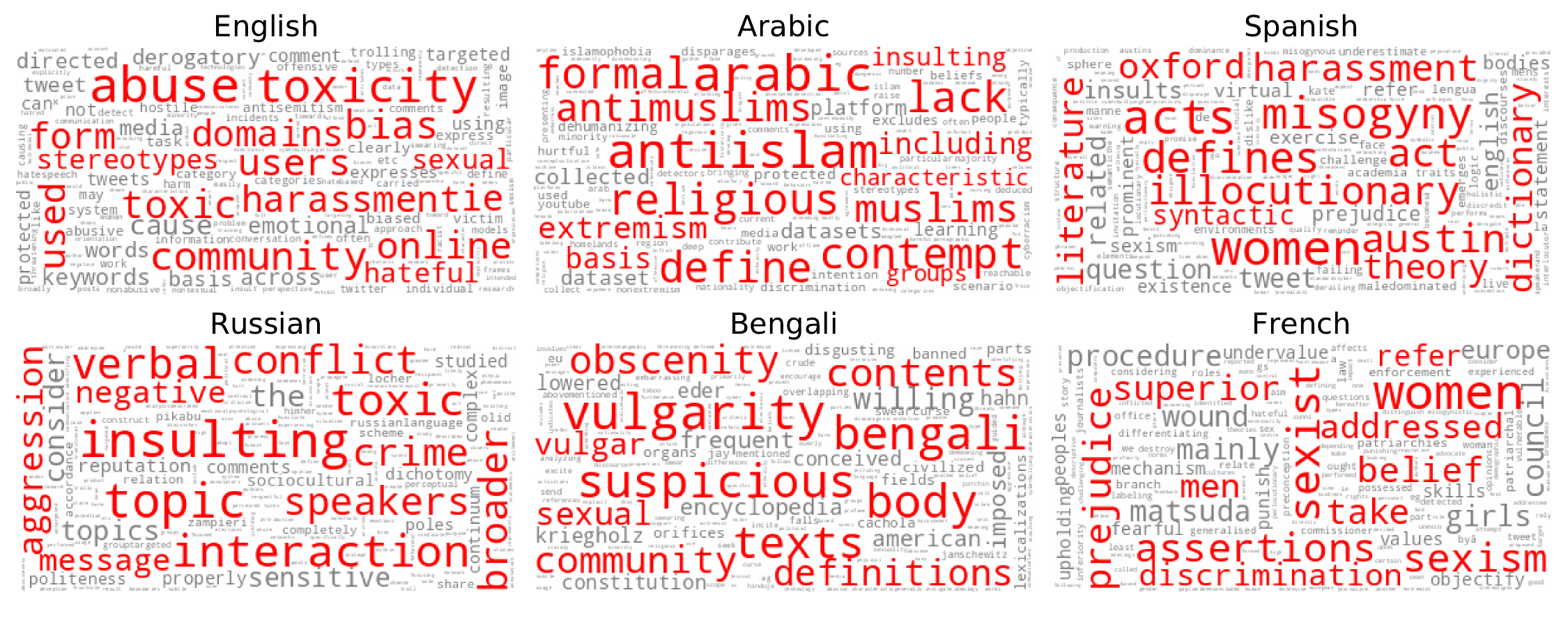}
    \caption{\textbf{Wordclouds summarizing the construct definitions across different languages, with the color highlighted by TF-IDF scores of the keywords.} While English and Bengali datasets' constructs are defined to include many different targets, we can see gender is emphasized in French and Spanish datasets, while race and religion are more pertinent in Arabic datasets.}
    \label{F9}
\end{figure*}

\section{Results}
We provide an overview of the state-of-the-art in hateful communication datasets, before presenting our findings on the identities that shape them. First, we analyze the location of the authors' institutional affiliations, the languages of the datasets they contribute, and the topics they focus on. Next, we study which targets of hateful communication are explicitly mentioned in the papers that introduce new hateful communication datasets, to map out how research has distributed its efforts across different target groups. Finally, we analyze the targets that are empirically included in the datasets to assess the possible mismatches between the conceptualization of hateful communication and the resources intended to address it.

\subsection{Summary of Collected Papers \& Datasets}

Out of 2533 papers initially matching our search queries, we identified 341 papers introducing novel datasets about hateful online communication suitable for training machine learning models. Figure \ref{F3} shows the increase in the number of datasets shared over time, spanning publication years 2012–2022.

\subsubsection{Dataset Multiplicity}
While most papers (261 out of 341) introduced one hateful communication dataset, 80 papers introduced two or more datasets. Overall, the research introduced 492 datasets on hateful communication.

\subsection{Q1: In what countries are producers of hateful communication datasets located, what languages are they studying and how are the datasets' qualities evolving?}
We now turn to the stakeholders of the hateful communication datasets. We start by unpacking the context of the production of the datasets. We investigate the diversity of researchers contributing to this body of work, using the location of their institutional affiliations to situate the researchers.
We show how researchers' locations are correlated with differences in the choice of languages and topics covered by the datasets they produce. We analyze the quality of hateful communication datasets with respect to both intrinsic data quality indicators---such as the diversity and coverage of languages, platforms, and targets---as well as extrinsic factors---that include the accessibility and interoperability of datasets.

\subsubsection{Dataset Availability}
51\% (251) of all datasets in our sample are publicly available (i.e., authors provide links to the dataset or specify that the data is available upon request; we also included those papers that are only available upon request since terms of use of social media platforms often hold authors back from sharing their data in public data repositories). Among those that are available, the most common way to distribute datasets is GitHub, with 62\% of datasets shared via Github, followed by some open repositories such as Zenodo (5\%) and Google Drive (2\%), and the rest are available through a provided link to websites. Only 7\% of datasets are specifically available upon request via email or any given contact.

Although all 180 publications corresponding to the 251 publicly available datasets mention ways to access the data, we found that it was not possible to access 21 datasets of 17 papers due to wrong or expired links. The remaining 163 out of 180 studies provided valid access to 230 novel datasets.

\subsubsection{Diversity of Data Sources}
Next, we turn to the platforms that researchers used as sources for data collection. Most of the publications collected the datasets on a single platform (252), while the remaining were from two (45) or more platforms (44) (Figure \ref{F4}). Twitter is the most popular data source, featured in 55\% (272) of all datasets, followed by YouTube (11\% (54) datasets), Facebook (10\% (50) datasets), Reddit (8\% (40) datasets), Wikipedia (7\% (33) datasets), and Instagram (3\% (16) datasets).

\subsubsection{Diversity of Researchers' Locations}
The United States is the country leading the production of hateful communication datasets. Overall, U.S.-based researchers were involved in the creation of 135 new datasets---27\% of the total number of datasets (492). In decreasing order, the remaining locations with the most contributed datasets are India (12\% (57) datasets), the United Kingdom (12\% (57) datasets), Germany (7\% (32) datasets), and Italy (7\% (32) datasets)(Figure \ref{F5}). 

Yet, the field appears changing over time. Not only is the field growing in size (as noted in the previous section and depicted in Figure \ref{F3}), but it is concurrently expanding its geographic borders. As of 2022, researchers with affiliations in 59 countries contributed new datasets, compared to 9 countries before 2017. Especially, since 2017, each of the top-5-dataset-producing countries contributed fewer datasets than the non-top-5 countries taken together (see Figure \ref{F6}a). Moreover, since 2017, transnational collaborations increased: 23\% of all datasets in our sample were published by transnational teams since 2017, while only 1\% of datasets were published by transnational teams before 2017. 

Also when accounting for the low number of datasets before 2017, we see a difference in the frequency of transnational collaborations: before 2017, only 4 out of 23 datasets (17\%) were created by transnational teams, while since 2017, 25\% of datasets fall into that category.

This highlights the increasing diversity of teams that contribute hateful datasets but also shows that large parts of the world are not involved despite experiencing hate online.

\begin{figure*}[h]
    \centering
    \includegraphics[width=\textwidth]{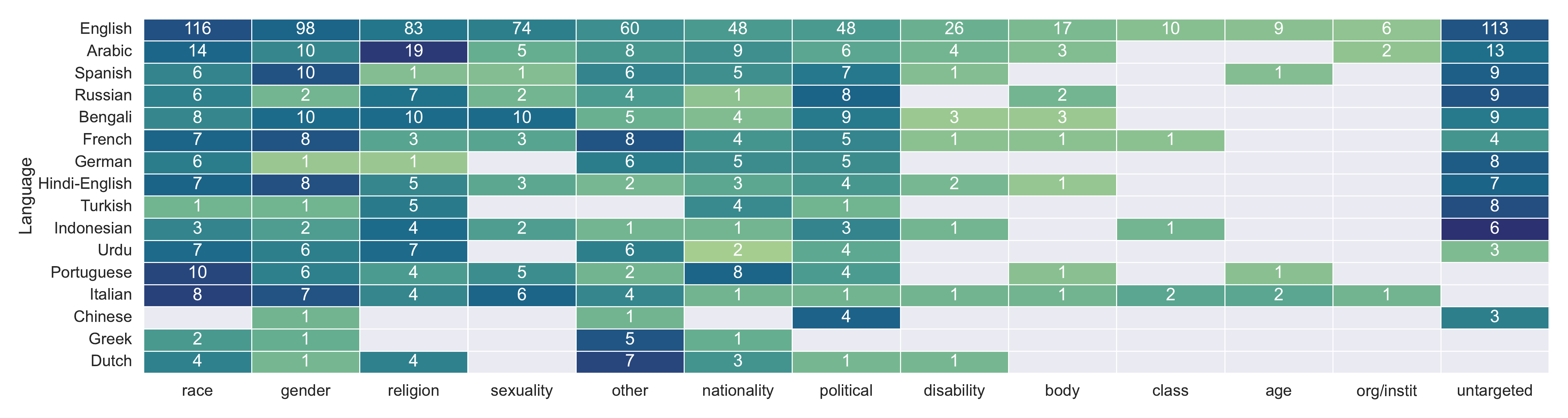}
    \caption{\textbf{Distribution of conceptual or operationalized targets across languages.} The most popular target categories are race, gender, and religion. Race is the most frequent target studied in English, German, Portuguese, and Italian datasets. In contrast, gender-based abuse is widely studied in Spanish, French, and Hindi-English code-mixed datasets, and religion is the most frequent target for Arabic, Turkish, and Indonesian, and one of the main targets in Bengali and Urdu. Other target attributes like class, disability, and age are rare.}
    \label{F10}
\end{figure*}

\subsubsection{Diversity of Dataset Languages}
The hateful communication datasets in our sample span over 49 languages, with English being the most common language (61\% (298) datasets), followed by Arabic (8\% (37) datasets), Spanish (6\% (29) datasets), Russian (5\% (25) datasets), Bengali (4\% (22) datasets),\footnote{Bengali datasets are also published under \textit{``Bangla"}, which is the endonym for Bengali.} French (4\% (22) datasets), German (4\% (20) datasets), and Hindi-English (3\% (17) datasets),\footnote{We merged all the datasets annotated as Code-mixed Hindi and a mixture of Hindi and English to Hindi-English.}. Some datasets also focus on low-resource languages such as Urdu (2\% (11) datasets), Kazakh (0.4\% (2) datasets), and Malay (0.2\% (1) dataset). 90\% of all datasets are monolingual (Figure~\ref{F7}), 5\% contain text in two languages, and 5\% include more than two languages. 

Figure~\ref{F6}b shows that the number of datasets in English has steadily risen since 2012. However, similarly, as for the locations of the researchers, we have witnessed a diversification of the languages since 2017/2018. 
Before 2017, almost all datasets were in English. After 2018 there are almost as many datasets in non-top-5-most-common languages as there are in English each year.

Next, we unpack how the language of the datasets relates to the location of the researchers authoring them. Figure~\ref{F8} shows the distribution of datasets by their language and the location of their authors. Researchers in selected locations, such as the U.S. and U.K., contribute to datasets in a range of languages. The converse appears also true: English is the most common dataset language even for researchers in countries that do not speak English as an official language. For all other languages, the majority of datasets are contributed by countries that speak the language itself officially, e.g., Bengali datasets mainly originate from Bangladesh and Russian datasets from Russia. Arguably, this may reflect the contextual nature of hateful communication, which requires not only linguistic proficiency, but also deep cultural situatedness---researchers' own experiences (historical, cultural, familial, and personal) shape the way they act in the world around them, in this case through their focus on different languages, and possibly different phenomena: e.g., what hateful communication looks like in Germany may be shaped by its current sociopolitical condition as well as its history, and therefore may significantly differ from hateful communication in neighbouring countries such as France. This opens questions on the ability of hateful communication research to be effective in social contexts that are currently not represented in datasets.

\subsubsection{Diversity of Topical Focus}
Hate speech is the most popular construct covered by research in hateful communication with 28\% (138 datasets from 97 publications), followed by cyberbullying with 13\% (62 datasets from 40 publications), offensiveness with 13\% (62 datasets from 39 publications), abusiveness with 12\% (59 datasets from 40 publications), toxicity with 8\% (38 datasets from 27 publications), and sexism with 3\% (13 datasets from 10 publications). 

Given the relevance of situatedness for hateful communication research, we further unpack the relationship between how researchers define the construct under study and the language of the datasets. 
We find that how researchers define a construct varies significantly depending on the language of the dataset. To gain qualitative insights into those differences we use word clouds that surface the most discriminative terms used in construct definitions across various languages. Figure~\ref{F9} shows the differences between the top 6 languages in our sample. To surface the most discriminative terms used in construct definitions across various languages, we use an approach based on Term Frequency Inverse Document Frequency (TF-IDF). Specifically, we combine the construct definitions of all datasets in a language into a single document, compute the TF-IDF scores of each keyword in all documents, and then highlight keywords with a TF-IDF above the threshold of 0.01. This approach highlights the words that are especially salient for certain languages, instead of words that are common across all languages. Our results show that gender is emphasized in French and Spanish datasets, while race and religion are more pertinent in Arabic datasets. This highlights the diversity of conceptualizations of hate across different languages. Considering those differences is especially important when researchers merge and translate datasets to train hateful content detection systems which is a promising approach, especially for under-resourced languages \citep{rottger-etal-2022-data}.

\subsubsection{Summary}
 Our results highlight that the quality of datasets according to extrinsic factors is relatively low. Only half of the datasets are directly accessible, and data interoperability may be hindered by discrepancies in how researchers conceptualize the construct. 
 However, we see an increase in multilingual datasets over time, which is a positive indicator for the dataset quality, since it signals increases in the diversity and coverage of the dataset.
Before 2018, U.S.-based researchers led the production of hateful communication datasets, which are predominantly in English. Since, with a growing number of researchers from various backgrounds involved, the overall volume and the diversity in languages and targets of hateful communication have increased, which also contributes to an improved representation of hateful communication in the collection of datasets in the field as a whole. 

\subsection{Q2: Which identities are discussed as targets of hateful communication in the scientific literature?}
The previous section analyzed the temporal evolution of the production of hateful communication datasets and unpacked how those identities relate to different practices that affect the qualities of the datasets. Next, we analyze the identities that this body of research focuses on---in particular, the identities of the targets of hateful communication. Following frameworks for assessing the quality of datasets \citep{sen2021total}, we track how targets are included in the conceptualization phase of the work---i.e., in the definitions of the constructs under study---in the operationalization phase---i.e., in the choice and design of automated or manual labeling procedures, as well as sampling and data collection procedures.

\subsubsection{Diversity of Targets in the Literature}
Here we show how the literature divides its effort among targets that are mentioned in the construct definition and/or are considered in the data creation process. In this analysis, we combine conceptualized and operationalized targets for the large sample of 341 papers.

\begin{figure}
    \centering
    \includegraphics[width=.45\textwidth]{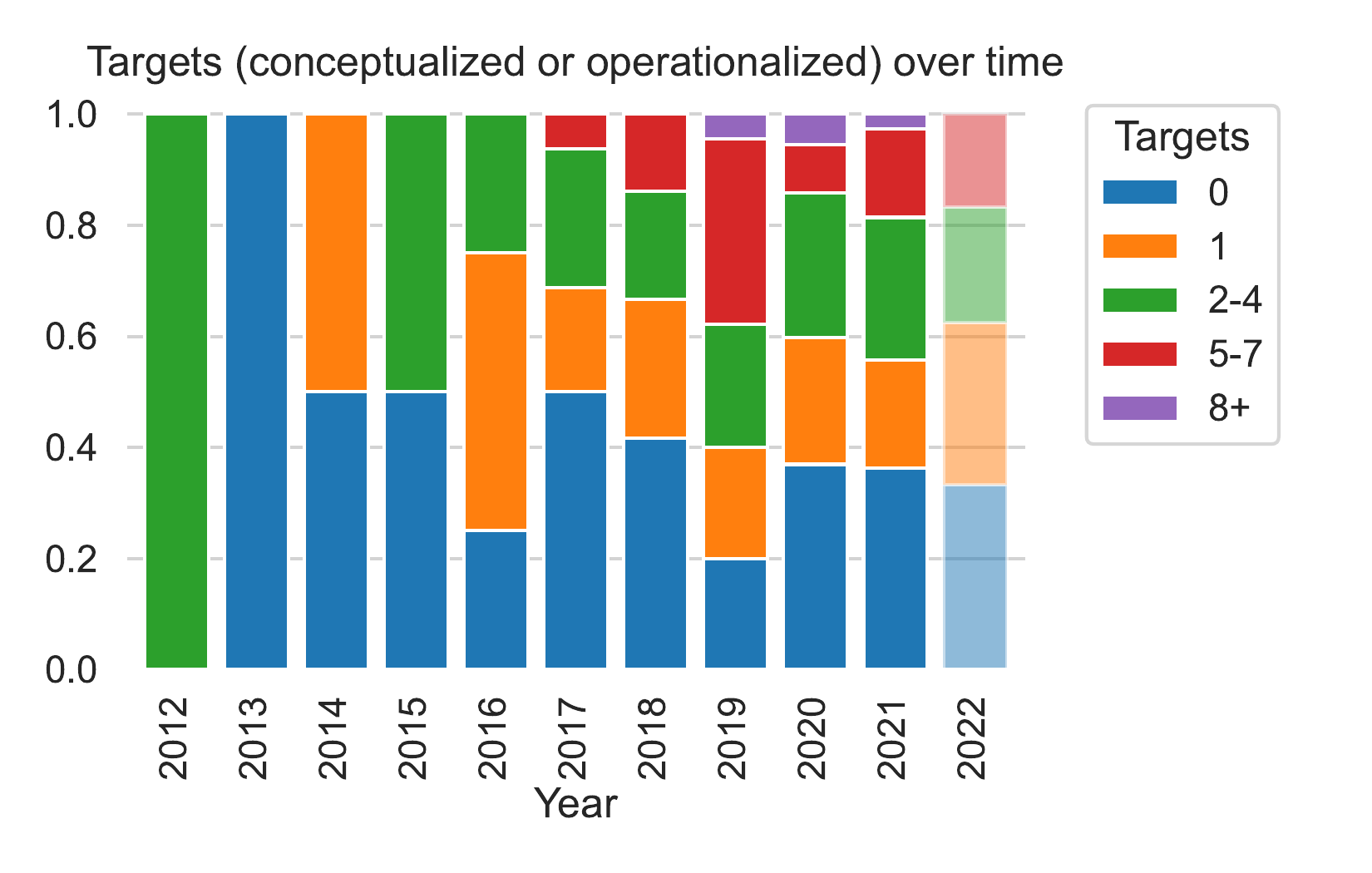}
    \caption{\textbf{Temporal variability of targets included in the literature.} We see that there has been a gradual rise towards multi-target datasets.}
    \label{F11}
\end{figure}

Figure \ref{F11} shows the distribution of papers that describe targets explicitly over time. Overall, 64\%  (314) of all datasets mention at least one specific target group, contrasted to the other 36\% (178) of datasets that explore hateful content as a general phenomenon without covering any identifiable target groups. 

\begin{figure}[h]
    \centering
    \includegraphics[width=0.45\textwidth]{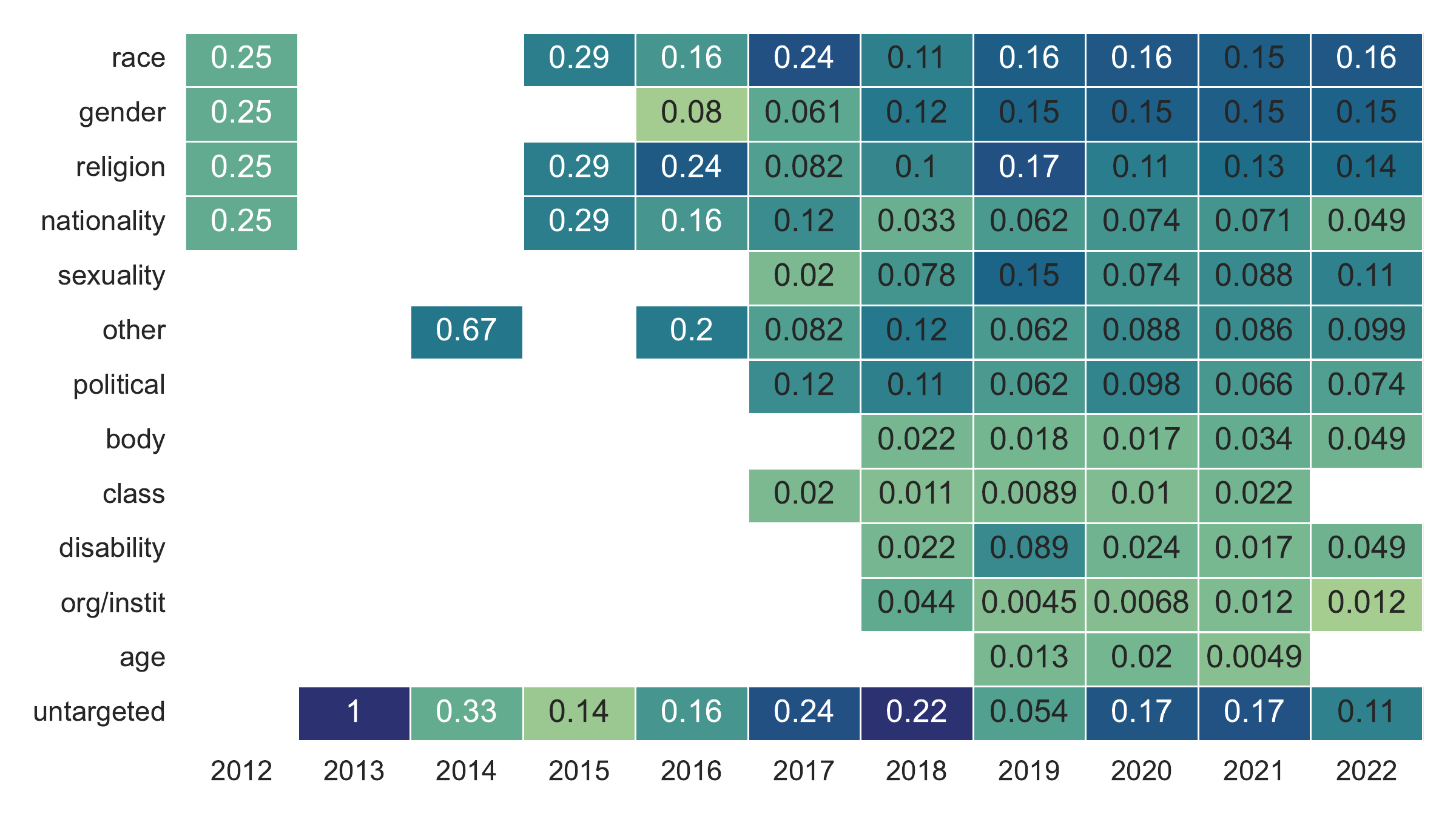}
    \caption{\textbf{Yearly proportion of targets in datasets over time based on our literature review.} There has been an increase in attention towards targets beyond race, gender, and religion, especially after 2017.} 
    \label{F12}
\end{figure}

The most common target identities are race (38\% (188) datasets), gender (35\% (172) datasets), religion (32\% (158) datasets), sexuality (22\% (110) datasets), and political affiliation (19\% (92) datasets). There are also 21\% (103) of all datasets containing targets outside our predefined categories which we label \textit{``other"}; examples of other targets include celebrities~\citep{lu2020cyberbullying} and students~\citep{del2014aggressive}. Although certain targets are underrepresented in hateful communication datasets (e.g., groups discriminated against for their age, body image, social class, and organizational/institutional affiliation), overall, this research space appears in the process of becoming more inclusive: since 2017, the number and diversity of targets increased compared to the preceding five years. Figure~\ref{F11} summarizes how the normalized distribution of targets shifted over time, whereas Figure~\ref{F12} shows the multiplicity of targets per dataset and year.

When looking at the diversity of targets that are represented in different languages (see Figure \ref{F10}), we see differences between languages that are also in line with the differences in the conceptualizations that we discussed before (see Figure \ref{F9}). For example, religious hatred, particularly Islamophobia is a major focus in Muslim-majority countries where Arabic, Turkish, Bengali, and Urdu are spoken (Turkey, Bangladesh, Pakistan, etc.), while gender is the most frequent target in Spanish, French, and Hindi-English datasets. While those variations in the prominence of targets may be a reflection of specific cultural and political factors in countries where these languages are spoken, it is important to consider those differences when datasets are used for training multilingual hate detection systems.

\subsubsection{Summary}
Datasets are increasingly specific about which targets they aim to include by applying more refined and targeted sampling strategies, and diverse in the range of targets they cover. However, datasets rarely cover certain target identities such as age, body image, and organizational/institutional affiliation. 
Differences in which targets are represented in different languages are very pronounced and can hinder the interoperability of datasets.

\begin{figure*}[h]
    \centering
    \includegraphics[width=\textwidth]{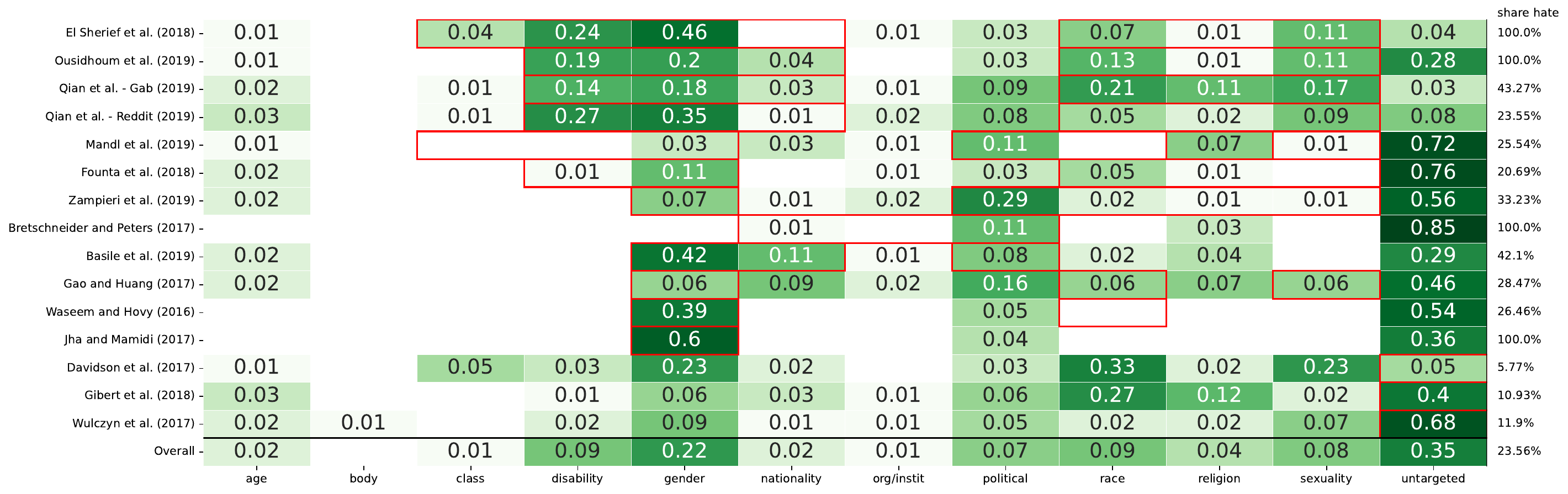}
    \caption{\textbf{Distribution of instances labeled as \textit{``hate"} over the target categories per dataset.} To the right of the heatmap, the share of instances labeled as \textit{``hate"} is indicated for each dataset. The last row shows the distribution over the target categories for all datasets aggregated. Darker shades of green correspond to a higher share of targets in the respective target category. A red frame around a cell indicates that the target category has been either explicitly conceptualized or operationalized by the dataset creators in the corresponding publication. If the publication does not explicitly conceptualize or operationalize any targets, \textit{``untargeted"} is highlighted.}
    \label{F13}
\end{figure*}

\subsection{Q3: Which identities are included as targets in hateful communication datasets, even if not explicitly mentioned in the literature? }

Next, we assess how the targets that are described in the literature are empirically included in the datasets themselves. First, we report on the diversity of dataset creation strategies observed in the literature, briefly discussing their potential impact on the composition of resulting datasets. Second, we aim to identify, if any, the discrepancy between conceptualized, operationalized, and detected targets. For this in-depth analysis, we focus on a convenience sample of 15 widely-cited and easily accessible English-language datasets. Three computational social science (CSS) researchers independently annotated conceptualized and operationalized targets in the 15 publications, discussing differences until consensus. 
Then, we computationally analyze the datasets to find detected targets, which may be present regardless of whether they were conceptualized or operationalized in the accompanying publications. 

We first give an overview of the different dataset creation strategies found in the literature, then compare conceptualized and operationalized targets, and finally compare them in aggregate against detected targets.

\subsubsection{Diversity and Impact of Dataset Creation Strategies} For 34.6\% of datasets (170 datasets from 131 publications), the procedure to collect the dataset is explicitly mentioned. The most common strategy is full or partial random sampling (34\% (57) datasets), followed by the use of a specifically developed lexicon, corpus, dictionary, or otherwise assembled list of topically relevant terms and keywords (31\% (52) datasets). Other than that, the focus on specific languages (8\% (14) datasets) and the use of classifiers (5\% (9) datasets) are additional dataset creation strategies in active use.     

There is a direct link between the dataset creation strategy used and the composition of the resulting dataset. Datasets created using fully or partially random sampling are expected to cover a broad range of targets (e.g., the dataset by \citealp{wulczyn2017ex}), while datasets that result from term- and keyword-based lists naturally tend to more precisely capture the specific group of targets operationalized via the underlying list (e.g., the dataset by \citealp{waseem2016hateful}).

 \subsubsection{Conceptualized vs. Operationalized Targets}

\begin{table}[h]
\small
    \caption{Confusion matrix, showing the (mis-)match between conceptualizations and operationalizations of targets. The (mis-)matches are analyzed on a dataset level based on our convenience sample.}
    \centering
    \begin{tabular}[t]{p{2.2cm}|p{2.2cm}|p{2.2cm}|}
          & Operationalized & Not \\
          & & Operationalized \\ \hline
         Conceptualized & 3 & 7 \\ \hline
         Not & & \\
         Conceptualized & 2 & 3 \\ \hline
    \end{tabular}
    \label{tab:confusion_targets}
\end{table}

Table \ref{tab:confusion_targets} reports the overall discrepancy between conceptualized and operationalized targets in the convenience sample datasets. 

The majority of datasets (7 out of 15) conceptualize targets without explicitly labeling which targets are present in their datasets. Upon close inspection, we find that targets are often included in the annotation instructions, but the annotators' task is ultimately to label whether a message contains hateful communication or not in an untargeted, binary way. We speculate that this is for a cost/benefit trade-off. Firstly, annotating targets can be expensive since it requires additional time and effort. Secondly, the downstream applications are often formulated as binary problems---for many benchmarks and shared tasks, models are expected to identify hateful communication and not necessarily its targets. 

2 out of 15 datasets operationalize more targets than they conceptualize. We find this is due to post hoc analyses where the authors of the datasets decided to label data characteristics they found interesting. As an example, ~\cite{bretschneider2017detecting} set out to study anti-immigrant hate, but after finding several discussions about politicians in the data, they included \textit{``politicians"} as one of their operationalized targets. Three datasets neither conceptualize nor operationalize targets because they study phenomena like general abuse or toxicity. Only three datasets operationalize the exact targets they conceptualize. This finding is surprising and stresses the need for standardized reporting practices in the field.

\subsubsection{Conceptualized/Operationalized vs. Detected Targets}
Considering the inconsistencies around conceptualized and operationalized targets observed in our convenience sample, we combine them before comparing them against detected targets. In the following analyses, we focus on the proportion of the datasets that are labeled as containing hateful communication because those instances supposedly include conceptualized and operationalized targets.

Figure \ref{F13} shows the distribution of detected targets. Black frames denote conceptualized/operationalized targets. Each cell displays the proportion of instances containing each detected target category. Instances for which we could not detect any target are reported in the \textit{``untargeted"} column.

Among those publications that explicitly conceptualize or operationalize specific targets, we find that the fraction for single categories of detected targets that were not part of these definitions ranges between 1\% and 16\%. Although relatively small, the presence of non-conceptualized targets is consequential, as it may impact the performance of classifiers trained on the data in terms of divergent validity---e.g., a classifier claiming to detect hate toward women (conceptualized target: gender) may in practice detect opposition to left-wing ideology (detected target: political affiliation), which may be empirically correlated in the dataset; the application of such classifier would have unintended consequences, such as censoring political views rather than preventing harm. 

While three papers in our sample did not declare targets in their conceptualization or operationalization we find that empirically, the corresponding datasets cover a wide range of detected targets in varying proportions (1\% to 33\% of the instances). This unequal distribution may negatively impact reliability in detecting hateful communication toward underrepresented targets. Note that the issue of distribution among different targets is also present in papers that conceptualize and operationalize targets. We provide suggestions for researchers on how to critically address this issue in the discussion section.

Finally, we identify an overarching pattern. Works that explicitly operationalize the targets they conceptualized \citep{jha2017does, waseem2017understanding, ousidhoum2019multilingual}, also include a higher fraction of matching detected targets---in other words, the datasets are empirically good fits with what was documented in the paper and annotated by labelers. In contrast, targets that are conceptualized but not operationalized are also frequently underrepresented as targets in the datasets, or missing altogether \citep{gao2017detecting,mandl2019overview,qian2019benchmark,founta2018large,basile2019semeval,zampieri2019predicting}. We speculate that including targets in the operalization phase prompts authors and labelers to be aware of them throughout the whole dataset creation process. Hence, what benefits from such a practice is not just the quality of documentation, but also the quality of the dataset itself.

\subsubsection{Summary}
Conceptualized and operationalized targets match in only 20\% of datasets in our sample. For datasets that conceptualize or operationalize targets, up to 16\% of their instances contain targets that were never conceptualized nor operationalized, which may make classifiers perform unpredictably on such targets. Datasets that do not declare targets at all cover targets unequally, which may impact the accuracy of classifiers on underrepresented targets. It is crucial to underscore that our findings are derived from a convenience subset of 15 datasets. While these datasets hold significant prominence within the domain of hateful communication, being frequently referenced sources, future research is needed to assess if the observed patterns persist in larger samples.

\section{Discussion \& Conclusions}

\subsection{The State of Hateful Communication Datasets}
This review proposed a broader outlook on the quality of hateful communication datasets, the identities of the targets of hate, as well as the linguistic diversity and backgrounds of the researchers involved in the processes of data collection, annotation, and curation.
 In the context of the computational study of hateful communication, we unpacked how the identity and diversity of the targets included in the research not only depend on the identities of the data annotators but also on the wider research context in which datasets are created. We found that the production of hateful communication datasets is concentrated in selected locations: researchers with affiliations in the U.S. contribute over one-fourth of the datasets. Similarly, almost two-thirds of the datasets are in English. This is in contrast with the situatedness of hateful communication research: languages other than English are mainly covered by researchers located in native-speaking countries, and the constructs that they study differ depending on such languages. This arguably reflects the need for deep knowledge about the context where hateful communication takes place.

Furthermore, it is important to discuss the potential causes and effects of this apparent conflict between an established U.S.- and English-centric mainstream, and the need for contextualization in the specific domain of study. On the one hand, the relative homogeneity of datasets cast doubt on their suitability to train machine learning models that capture context-specific aspects of hateful communication. We find that promisingly, the production of hateful communication datasets is not only growing in output but also diversifying in who is represented in it—in terms of both the researchers that produce it and the identities that are included in the data. 

On the other hand, multiple factors may slow down the diversification of this field. We argue that researchers may be incentivized to produce datasets in English both explicitly and implicitly. Producing datasets is costly: obtaining reliable annotations for a wide range of targets and at a scale requires substantial investment. Moreover, curating datasets in new languages may require developing specialized resources, contrasted with highly available tools and techniques to sample and process data in English. Therefore, it is unsurprising that new datasets rarely introduce new languages, and that the Global North is among the largest producers of datasets, which raises concerns about the ability of this field to avoid reproducing inequality. Furthermore, curating datasets in English may widen the user base for the datasets and consequently, increase the visibility of the research. 

In this light, institutionalized incentives for promoting local impact are essential to sustain the diversification of the field, especially through the inclusion of a broader range of researchers.

\subsection{Implications for the Users of Hateful Communication Datasets}
Although research is broadening its attention towards a diverse range of targets of hateful communication, we found that some identities (such as class, disability, and age) receive less attention. Moreover, a sizable fraction of the publications to date do not specify which targets are included in their datasets, and for the publications that do, there is a discrepancy between the targets documented in the publications and those effectively present in the datasets. These findings can be problematic for the users of hateful communication datasets---especially those who train machine learning models on such datasets to detect hateful communication in different application domains.

Lack of representation, under-representation, and undocumented representation of targets all make machine learning models unable to perform accurately and predictably. Determining an ideal ratio of targets in a dataset may not be practical or possible. However, for research to have a positive real-world impact, the dataset should precisely represent the targets it aims to capture. Future research could explore participatory approaches \citep{maronikolakis2022listening}, data documentation \citep{miceli2021documenting}, and theory-informed and target-aware data collection procedures \citep{li2021target, samory2021call, uyheng2022language} as promising avenues to overcome and document the mismatch between the conceptualized, operationalized, and detected targets.

\subsection{Implications for the Creators of Future Hateful Communication Datasets}

We highlighted a gap in how present-day datasets cover the variety of hateful communication targets. Especially, the field is moving towards including multiple targets in each dataset, which enables sophisticated computational modelling and robust evaluation. Thus, there are ample opportunities for the creation of novel datasets that better serve minority identities. Furthermore, we find a shortage in studies focusing on intersectionality---while some papers do explicitly operationalize intersectional targets~\citep{waseem2017understanding,vidgen2020directions}, they are a small portion of datasets studied in this work.  

Yet, we also identified avenues for improving data practices, to sustain high-quality standards in this socially-relevant field. We stress the importance of clearly including targets in all phases of the dataset creation process, starting with the conceptualization of the construct that the dataset aims to capture (whereas, several papers to date omit this crucial information). Targets should also be included in the operalization phase of the data curation process, e.g., in sampling strategies, annotation instructions, and annotation labels if possible (whereas, almost half of the conceptualized targets were not operationalized in the literature). Finally, authors should take steps to identify targets in their collected dataset that they did not explicitly intend to include, and be upfront about their treatment of such cases (whereas, we found that almost all papers in our sample had targets included in their dataset that were not described in the paper). 
Overall, we surfaced the need for better conceptualization, operationalization, and documentation practices around targets. To this end, we believe it a fruitful avenue of research to develop tools and standard procedures to aid the fine-grained documentation of targets.

\subsection{Future Work and Limitations}

This review proposes ways forward to improve the quality of datasets in future studies. First, instituting benchmarks, measures and shared tasks to empirically evaluate the dataset generalizability across contexts and targets may promote critical approaches to data quality; Second, research on the science of computational social science should arguably aim for improved measures of data quality, to establish causal links between the quality of datasets and the practices, views and characteristics of those involved in the creation. Finally, we argue that higher reflexivity in this research field may be beneficial, such as via positionality statements in papers and datasets. 

The study presented has some limitations that must be considered when interpreting its results. Although we attempted to cover most of the existing literature and datasets on hateful content, we made some restrictive assumptions that may have resulted in missing works. Firstly, we only examined three academic databases, which may not have included all relevant publications. Secondly, we only included English publications, meaning that language-specific conferences were excluded, potentially leading to the exclusion of relevant research. Lastly, we excluded all papers that did not mention whether a newly created dataset was introduced based only on the information provided in the abstract. This may have led to an increased number of false-negative decisions.

Furthermore, we discovered during the annotation process that while it is relatively easy to decide if a paper talks about targets or not, it is often non-trivial to identify the conceptualization of targets and differentiate it from the operationalization. For example, in \cite{warner-hirschberg-2012-detecting}, hate speech and potential targets are mentioned in several sections of the paper, such as the introduction and related work. Therefore it is not easy for annotators of the literature review to determine the target conceptualization employed in the context of the respective work. On the other hand, based on the annotator instructions, it is clear that targets are explicitly annotated as \textit{anti-semitic, anti-black, anti-Asian, anti-woman, anti-muslim, anti-immigrant, or other-hate} \citep{warner-hirschberg-2012-detecting}. To avoid building our analysis on annotations where we know that annotators had a hard time and also in part disagreed, we decided to merge the conceptualized and operationalized target annotations for the large sample. For the small sample, we provide an in-depth discussion of the differences together with more detailed and time-consuming annotations. 

Also, our dictionary-based approach to identifying targets is limited and prior work suggests that the Hatebase lexicon includes terms that are generally not used in a hateful context \citep{davidson2017automated}. Our small validation study does not assess the recall of the dictionary but shows that the precision is acceptable (0.68).  We apply our dictionary only to the small sample and acknowledge that it would have been ideal to conduct this analysis on the complete body of datasets. However, we are confident that the subset provided by \citet{risch2021toxic} is an adequate representation.

Finally, our study of targets is abstracted at a demographic identity level, focusing on broad categories of race, gender, religion, etc. We do not distinguish between finer-grained identities within these, e.g., we do not differentiate Islamophobia from anti-Christian rhetoric. Future research can build on our work to specifically measure the representation of persecuted minority groups within these categories.

\section{Data Availability}
We acknowledge the importance of open and transparent data sharing and adhere to the FAIR (Findable, Accessible, Interoperable, and Reusable) principles. The data used in this study are available via a data archive for non-commercial purposes, subject to any necessary ethical and legal considerations. 
\footnote{The fully annotated table of articles and scripts for figures are available on \url{https://github.com/uzeui/HateComm\_Review.git}.}

\section{Funding}
This work was created in the context of the project: Digital Dehumanization: Measurements, Exposure and Prevalence (DeHum), funded by the Leibniz Association Competition (P101/2020).

\clearpage

\appendix
\begin{appendices}

\subsection{Further Details on the Search and Selection Procedures}

\subsubsection{Search Procedure}
The queries used for retrieving publications from the different literature search engines are provided in Table \ref{tab:appendix}.

%Table for the search query
%\section{Appendix B. Literature Search Queries}
%\clearpage
\begin{table*}[t]\centering
    \begin{tabular}{l|c|p{10cm}} 
    \hline
    Search engines & Dimensions & Keywords \\
    \hline
    \multirow{3}{3cm}{\makecell[l]{Scopus\\ACL Anthology\\ACM Digital Library}} 
    & Topic & (dehuman* OR toxic* OR harass* OR (offensiv* OR offence) OR abusive* OR (hate* OR hating) OR bully* OR aggress* OR prejudi* OR troll* OR ``personal attack") \\  
    & Content & AND (speech* OR language* OR lingu* OR behavio* OR messag* OR content* OR text* OR discuss* OR conversation* OR tweet*) \\ 
    & Dataset & AND (dataset* OR corpus OR corpora) \\ 
    \hline
    \makecell[l]{ACL Anthology\\ACM Digital Library} 
    & Source/ Platform 
    & AND (``Social media" OR (online OR on-line) OR internet OR web OR ``social network" OR cyber OR digital OR forum) \\
    \hline
    \end{tabular}
    \caption{\textbf{Literature Search Engines and Queries.} We used the same search queries for three databases. Only for Scopus, we had to exclude the data source part of the query since it increased the number of papers drastically rather than filtering it down as expected.}
    \label{tab:appendix}
\end{table*}

After removing duplicates from the list of downloaded publications from the three search engines, we compared the resulting papers with dataset papers included in \url{hatespeechdata.com}, which is a dedicated website for cataloguing datasets annotated for hate speech, online abuse, and offensive language, with a list of datasets and keywords provided \citep{vidgen2020directions}. Almost all publications from \url{hatespeechdata.com} were included in our sample. We manually added the three missing publications to our sample (see Figure~\ref{F1}). %final sample. % but not in our first sample. 
Consequently, our final sample is a superset of  \url{hatespeechdata.com} and is 259\% larger than the \url{hatespeechdata.com} repository. 

\subsubsection{Selection Procedure}
 All publications in our final sample were manually screened by three experienced scholars with backgrounds in CSS and processed according to the inclusion and exclusion criteria described below. The reviewer training was separated into two parts, one for the title and abstract screening procedure, and the other for the full-text annotation procedure. Paper screening procedures were piloted among the experts before each round (see Figure \ref{F1}), and results were reviewed by the other authors independently for a sufficient inter-screener agreement level (Cohen’s kappa $>0.8$).

Discrepancies were discussed among the experts, and the conflicts left were discussed with other authors altogether to reach a mutual agreement. 

The first round of the screening was conducted by reviewing the titles of the publications to determine whether they met one of the following three inclusion or exclusion criteria:

\begin{enumerate}
\item[(i)]\emph{Topical Relevance}: The study should cover research on hateful online communication (e.g., explicit hate speech towards religion, as well as subtle and implicit forms of harm such as benevolent sexism, among others). Studies on misinformation, spam, fake news, or deception should be excluded. 

\item[(ii)]\emph{Dataset Novelty}: At least one dataset should be newly developed or adapted for the study consisting of text data or multimodal data where text is included, and the data needs to be collected from online sources. Papers using public datasets without any adaptations (e.g., new annotations or other expansions) should be excluded.

\item[(iii)]\emph{Computational Application}: Since the review focuses on datasets that can be used to develop computational methods to detect hateful communication online, we included only papers that mention computational methods such as specific machine learning or natural language processing methods. We exclude articles that focus only on manual data analysis of small samples of online data or data from other social science methods such as survey data, experiments, content analysis, or qualitative analysis.
\end{enumerate}

If it was unclear whether one of the criteria applied based solely on the title information, the papers were considered ambiguous and forwarded to the second selection step. In the second selection round, we screened the information provided in the abstracts of publications identified from the previous title review procedure (see Figure \ref{F1}). They had to meet all three of the inclusion or exclusion criteria above. The same procedure was applied to the inter-screener agreement check among the screeners, and a mutual agreement on the discrepancies with other authors was reached.

\begin{figure*}[h]
    \centering
    \includegraphics[width= 0.7\textwidth]{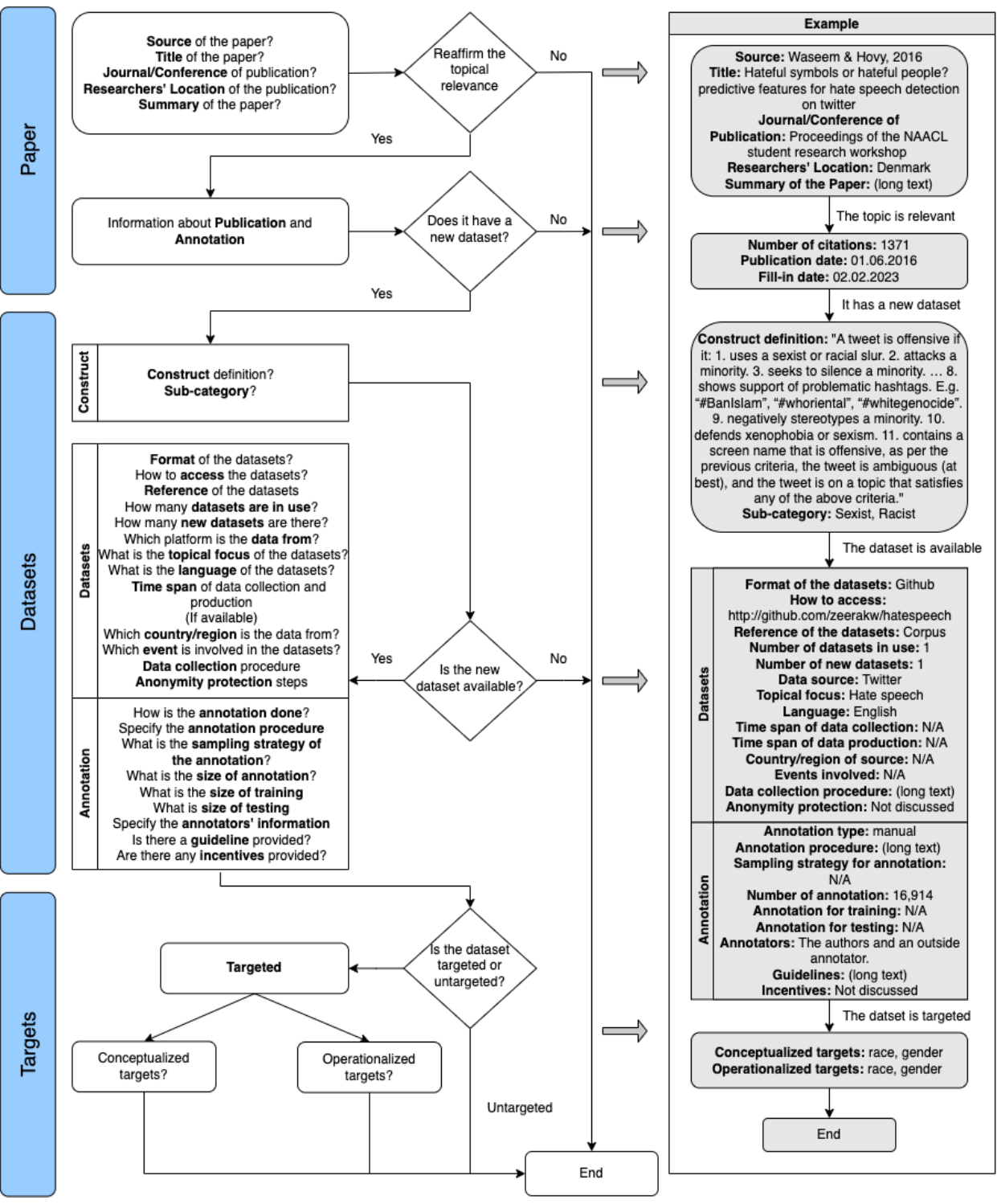} 
    \caption{\textbf{The activity diagram of literature survey and annotation with an example.}} 
    \label{F14}
\end{figure*}

\begin{table*}[t]\centering
    \begin{tabular}{|l|p{16cm}|}
    \hline
        Age & 60sfolks, babies, baby, boomerdeath, boomerdoomer, boomerentoomber, boomermoober, boomerremover, corbid, coronachan, deletus, doubledowndonnie, entomber, gaslighter, headassery, hitler, immunocompromised, karen, kid, kids, komekko, michiganders, old folks, oldaf, seniors, socialistremover, thankyouboomer, thintheherd, veterans, yoof, young folks \\\hline
        Body & beetus, condishuns, fatass, fatasses, fatlogic, fattie, fatties, fatty, fupa, gluttony, hambeast, hambeasts, hamplanet, hamplanets, hams, hog body, lard, obeast, obeasts, pig, piggy, whale \\\hline
        Class & bitter clingers, bludger, bog irish, bog trotter, bog trotters, bogan, bogans, booner, booners, carrot snappers, cave nigger, charva, charvas, chav, chavette, chavs, cracker, culchie, divvies, dole bludger, dole bludgers, gopniks, hayseed, hayseeds, hoodrat, hoodrats, hoser, hosers, jackeen, jant, jants, jhants, mokes, redneck, rednecks, rubes, scags, scallie, scallies, scally, scobe, sheepfucker, skanger, smick, spide, spides, spiv, spivs, stump jumper, trailer park trash, trailer trash, tree jumper, white trash, working class, yardie, yobbo, yobbos, yobs\\\hline
        Disability & cripple, cripples, gimp, gimps, gimpy, libtard, mongs, retard, retarded, retardeds, retards, spaz, sperg, spergs, tard, tarded, tards\\\hline
        Gender & bimbo, bint, bints, bitch, bitches, butt pirate, butt pirates, cock tease, cunt, cunts, degette, dyke, dykes, female, females, feminazi, fudgepacker, fudgepackers, girl, girls, himbo, hoe, hoes, hoodrat, hoodrats, hos, pussies, pussy, scags, shemale, shemales, slut, sluts, trannies, tranny, trans, trans people, trans women, twat, twats, whore, whores, woman, women\\\hline
        Nationality & african, africans, amerikkkan, banana bender, blaxican, bog trotter, bog trotters, bogan, bohunk, bohunks, booner, booners, camel fucker, camel fuckers, chinaman, chinamen, chink, chinks, chonky, dink, dinks, dogans, eyetie, fenian, fenians, flips, ginzo, globalists, greaseballs, greasers, guala, guido, guidos, guineas, heinies, hoser, hosers, hunyak, illegal, illegal alien, illegal aliens, illegal immigrant, illegal immigrants, illegals, indian, indians, jackeen, jap, japs, khazar, knacker, kraut, krauts, leb, limey, limeys, masshole, massholes, merkins, mungs, nips, paddy, paki, pakiland, piker, pikers, pikey, pikeys, plastic paddies, plastic paddy, polack, polacks, pommie, pommy, sand nigger, sand niggers, shanty irish, sheepfucker, sideways pussy, sideways vagina, sideways vaginas, smick, soup takers, surrender monkey, wexican, whigger, whiggers, wigga, wiggas, wigger, wiggers, wog, wogs, wop, wops, yardie, zionazi, zionazis\\\hline
        Org. / Instit. & cnn, cop, cops, female officer, female officers, law enforcement, media, police officer\\\hline
        Political & antifa, bolsofans, bolsonazi, ccp, commie, commies, commies, communist, communist, communists, conservative, conservatives, controlled opposition, cuckservative, damned party, deep state, dem, democrat, democratic party, democrats, dems, devilish leftists, dumb woke people, fascist, feminist, feminists, fucking liberals, fucking socialist, government, hellish, ideological garbage, iphone-stan socialist, lib land, liberal, liberals, liberals, libtard, libtard, libtards, male feminist, male feminists, milico, minions, mollusk, nazi, nazifascist, nazis, oval office, poor democrat, potus, president, prime minister, pt gang, racist, racists, red ideologist, republican, republican party, republicans, right winger, right wingers, scammer, socialist, socialists, white house, white supremacist, white supremacists\\\hline
        Race & alligator bait, anchor baby, ape, apes, beaner, bix nood, black people, blacks, blaxican, bogan, boonie, border hopper, border jumper, border jumpers, brown people, camel fucker, camel humper, camel humpers, camel jockey, camel jockeys, christ killer, coloreds, coloureds, congoid, coolie, coon, coon ass, cotton picker, cotton pickers, cowboy killer, cowboy killers, cracker, curry munchers, cushite, darkie, dindu, dink, dinks, dune coon, eurotrash, gook, gooky, groid, guala, gypo, gyppo, honkey, honkie, hoodrat, hoodrats, house nigger, hymie, injuns, jackeen, jap, jigaboo, jiggaboo, jiggaboos, jungle bunny, khazar, kikes, knacker, kneegroes, kneegrow, kushite, kyke, mokes, monkey, monkeys, moon cricket, moon crickets, mud duck, mud shark, neechee, negro, negroes, negros, nicca, niccas, nig, nig nog, nig nogs, nigers, nigga, niggah, niggahs, niggar, niggars, niggas, niggaz, nigger, niggerette, niggers, nigglet, nigglets, niggors, niggress, nigguh, nigguhs, niggur, niggurs, niglet, niglets, niglette, nigs, non white, non whites, nonwhite, nonwhites, octaroon, paddy, palefaces, peckerwood, peckerwoods, pickaninny, piker, pikers, pikey, plastic paddy, porch monkey, porch monkeys, quadroon, race traitor, raghead, ragheads, red bone, redskin, sand monkeys, sand nigger, sand niggers, savage, savages, shanty irish, sheboon, shit heel, sideways vaginas, slant eye, slant eyes, smoke jumpers, spear chucker, spic, spice nigger, spice niggers, spics, squaw, tar babies, tar baby, towel head, towel heads, trailer park trash, trailer trash, uncle tom, uncle toms, wagon burners, waspy, wetback, wetbacks, whigga, whigger, whiggers, white people, white trash, whites, whitey, wigga, wiggas, wigger, wogs, yellow bone, yellow bones, yids, yokel, yokels, zionazi, zipperheads\\\hline
        Religion & allah, camel fucker, camel fuckers, carrot snappers, christ killer, christ killers, christian, christians, clamheads, derka derka, dogans, durka durka, fenian, fenians, gerudos, gews, globalists, hebes, hebro, hebros, heeb, heebs, higger, higgers, holohoax, hooknose, hooknosed, hooknoses, hymie, islam, islamic, jew, jewish, jews, jihadi, khazar, khazars, kike, mackerel snappers, musla, muslamic, muslim, muslims, mussie, mussies, muzzie, muzzies, muzzrat, muzzrats, muzzy, orangies, oven dodger, oven dodgers, papist, papists, pisslam, proddy dog, proddy dogs, sand monkey, sand monkeys, soup taker , soup takers, synagogue satan, zio, ziojews, zionazi, zionazis, zionism, zionist\\\hline
        Sexuality & batiman, batty boy, batty boys, batty bwoy, batty bwoys, batty man, battyman, battymen, bull dyke, bull dykes, butt pirate, butt pirates, chi chi bwoy, closet fag, dyke, dykes, fag, fagbag, fagbags, faggot, faggots, faggy, fags, fudgepacker, fudgepackers, gay, gay mafia, gaylord, gaylords, gays, gender benders, ghey, gheys, homo, homos, lgbt community, lipstick lesbian, lipstick lesbians, pedophile, pedophiles, queer, queers, shemale, shemales, tranny\\\hline
        
    \end{tabular}
    \caption{\textbf{Target dictionary.} Target categories are presented with their respective keywords.}
    \label{tab:dictionary}
\end{table*}

\end{appendices}

\begin{thebibliography}{66}
\providecommand{\natexlab}[1]{#1}
\providecommand{\url}[1]{\texttt{#1}}
\providecommand{\urlprefix}{URL }
\expandafter\ifx\csname urlstyle\endcsname\relax
  \providecommand{\doi}[1]{DOI:\discretionary{}{}{}#1}\else
  \providecommand{\doi}{DOI:\discretionary{}{}{}\begingroup \urlstyle{rm}\Url}\fi

\bibitem[{Albadi et~al.(2018)Albadi, Kurdi and Mishra}]{albadi2018they}
Albadi N, Kurdi M and Mishra S (2018) Are they our brothers? analysis and detection of religious hate speech in the arabic twittersphere.
\newblock In: \emph{2018 IEEE/ACM International Conference on Advances in Social Networks Analysis and Mining (ASONAM)}. IEEE, pp. 69--76.

\bibitem[{Ayo et~al.(2020)Ayo, Folorunso, Ibharalu and Osinuga}]{ayo2020machine}
Ayo FE, Folorunso O, Ibharalu FT and Osinuga IA (2020) Machine learning techniques for hate speech classification of twitter data: State-of-the-art, future challenges and research directions.
\newblock \emph{Computer Science Review} 38: 100311.

\bibitem[{Baheti et~al.(2021)Baheti, Sap, Ritter and Riedl}]{baheti2021just}
Baheti A, Sap M, Ritter A and Riedl M (2021) Just say no: Analyzing the stance of neural dialogue generation in offensive contexts.
\newblock In: \emph{Proceedings of the 2021 Conference on Empirical Methods in Natural Language Processing}. pp. 4846--4862.

\bibitem[{Basile et~al.(2019)Basile, Bosco, Fersini, Nozza, Patti, Pardo, Rosso and Sanguinetti}]{basile2019semeval}
Basile V, Bosco C, Fersini E, Nozza D, Patti V, Pardo FMR, Rosso P and Sanguinetti M (2019) Semeval-2019 task 5: Multilingual detection of hate speech against immigrants and women in twitter.
\newblock In: \emph{Proceedings of the 13th international workshop on semantic evaluation}. pp. 54--63.

\bibitem[{Bretschneider and Peters(2017)}]{bretschneider2017detecting}
Bretschneider U and Peters R (2017) Detecting offensive statements towards foreigners in social media.
\newblock In: \emph{Proceedings of the 50th Hawaii International Conference on System Sciences}. pp. 2213--2222.

\bibitem[{Curtis(2023)}]{Curtis2023-ft}
Curtis WM (2023) Encyclopedia Britannica.
\newblock https://www.britannica.com/topic/hate-speech.
\newblock 2023-11-15.

\bibitem[{Daikeler et~al.(2023)Daikeler, Froehling, Sen, Birkenmaier, Gummer, Schwalbach, Silber, Wei{\ss}, Weller and Lechner}]{daikeler2023assessing}
Daikeler J, Froehling L, Sen I, Birkenmaier L, Gummer T, Schwalbach J, Silber H, Wei{\ss} B, Weller K and Lechner CM (2023) Assessing data quality in the age of digital social research: A systematic review.
\newblock \doi{10.31235/osf.io/gfbjk}.
\newblock \urlprefix\url{osf.io/preprints/socarxiv/gfbjk}.

\bibitem[{Davidson et~al.(2017)Davidson, Warmsley, Macy and Weber}]{davidson2017automated}
Davidson T, Warmsley D, Macy M and Weber I (2017) Automated hate speech detection and the problem of offensive language.
\newblock In: \emph{Proceedings of the international AAAI conference on web and social media}, volume~11. pp. 512--515.

\bibitem[{de~Gibert et~al.(2018)de~Gibert, P{\'e}rez, Garc{\'\i}a-Pablos and Cuadros}]{degibert2018hate}
de~Gibert O, P{\'e}rez N, Garc{\'\i}a-Pablos A and Cuadros M (2018) Hate speech dataset from a white supremacy forum.
\newblock In: \emph{Proceedings of the 2nd Workshop on Abusive Language Online (ALW2)}. pp. 11--20.

\bibitem[{Del~Bosque and Garza(2014)}]{del2014aggressive}
Del~Bosque LP and Garza SE (2014) Aggressive text detection for cyberbullying.
\newblock In: \emph{Human-Inspired Computing and Its Applications: 13th Mexican International Conference on Artificial Intelligence, MICAI 2014, Tuxtla Guti{\'e}rrez, Mexico, November 16-22, 2014. Proceedings, Part I 13}. Springer, pp. 221--232.

\bibitem[{ElSherief et~al.(2018{\natexlab{a}})ElSherief, Kulkarni, Nguyen, Wang and Belding}]{elsherief2018hate}
ElSherief M, Kulkarni V, Nguyen D, Wang WY and Belding E (2018{\natexlab{a}}) Hate lingo: A target-based linguistic analysis of hate speech in social media.
\newblock In: \emph{Proceedings of the international AAAI conference on web and social media}. pp. 42--51.
\newblock \urlprefix\url{https://doi.org/10.1609/icwsm.v12i1.15041}.

\bibitem[{ElSherief et~al.(2018{\natexlab{b}})ElSherief, Nilizadeh, Nguyen, Vigna and Belding}]{elsherief2018peer}
ElSherief M, Nilizadeh S, Nguyen D, Vigna G and Belding E (2018{\natexlab{b}}) Peer to peer hate: Hate speech instigators and their targets.
\newblock In: \emph{Proceedings of the International AAAI Conference on Web and Social Media}, volume~12. pp. 52--61.

\bibitem[{Fortuna and Nunes(2018)}]{10.1145/3232676}
Fortuna P and Nunes S (2018) A survey on automatic detection of hate speech in text.
\newblock \emph{ACM Comput. Surv.} 51(4).
\newblock \doi{10.1145/3232676}.
\newblock \urlprefix\url{https://doi.org/10.1145/3232676}.

\bibitem[{Founta et~al.(2018)Founta, Djouvas, Chatzakou, Leontiadis, Blackburn, Stringhini, Vakali, Sirivianos and Kourtellis}]{founta2018large}
Founta A, Djouvas C, Chatzakou D, Leontiadis I, Blackburn J, Stringhini G, Vakali A, Sirivianos M and Kourtellis N (2018) Large scale crowdsourcing and characterization of twitter abusive behavior.
\newblock In: \emph{Proceedings of the international AAAI conference on web and social media}, volume~12. pp. 481--500.

\bibitem[{Gao and Huang(2017)}]{gao2017detecting}
Gao L and Huang R (2017) Detecting online hate speech using context aware models.
\newblock In: \emph{Proceedings of the International Conference Recent Advances in Natural Language Processing, RANLP 2017}. pp. 260--266.

\bibitem[{Geiger et~al.(2021)Geiger, Cope, Ip, Lotosh, Shah, Weng and Tang}]{Geiger2021}
Geiger RS, Cope D, Ip J, Lotosh M, Shah A, Weng J and Tang R (2021) {"Garbage in, garbage out" revisited: What do machine learning application papers report about human-labeled training data?}
\newblock \emph{Quantitative Science Studies} 2(3): 795--827.
\newblock \doi{10.1162/qss_a_00144}.
\newblock \urlprefix\url{https://doi.org/10.1162/qss\_a\_00144}.

\bibitem[{Ghosh et~al.(2022)Ghosh, Ekbal, Bhattacharyya, Saha, Kumar and Srivastava}]{ghosh2022sehc}
Ghosh S, Ekbal A, Bhattacharyya P, Saha T, Kumar A and Srivastava S (2022) Sehc: A benchmark setup to identify online hate speech in english.
\newblock \emph{IEEE Transactions on Computational Social Systems} .

\bibitem[{Hada et~al.(2021)Hada, Sudhir, Mishra, Yannakoudakis, Mohammad and Shutova}]{hada-etal-2021-ruddit}
Hada R, Sudhir S, Mishra P, Yannakoudakis H, Mohammad SM and Shutova E (2021) Ruddit: {N}orms of offensiveness for {E}nglish {R}eddit comments.
\newblock In: Zong C, Xia F, Li W and Navigli R (eds.) \emph{Proceedings of the 59th Annual Meeting of the Association for Computational Linguistics and the 11th International Joint Conference on Natural Language Processing (Volume 1: Long Papers)}. Online: Association for Computational Linguistics, pp. 2700--2717.
\newblock \doi{10.18653/v1/2021.acl-long.210}.
\newblock \urlprefix\url{https://aclanthology.org/2021.acl-long.210}.

\bibitem[{Hardaker and McGlashan(2016)}]{hardaker2016real}
Hardaker C and McGlashan M (2016) "real men don’t hate women": Twitter rape threats and group identity.
\newblock \emph{Journal of Pragmatics} 91: 80--93.

\bibitem[{Jahan and Oussalah(2021)}]{jahan2021systematic}
Jahan MS and Oussalah M (2021) A systematic review of hate speech automatic detection using natural language processing.
\newblock \emph{arXiv preprint arXiv:2106.00742} .

\bibitem[{Jain et~al.(2020)Jain, Patel, Nagalapatti, Gupta, Mehta, Guttula, Mujumdar, Afzal, Sharma~Mittal and Munigala}]{Abhinav2020}
Jain A, Patel H, Nagalapatti L, Gupta N, Mehta S, Guttula S, Mujumdar S, Afzal S, Sharma~Mittal R and Munigala V (2020) Overview and importance of data quality for machine learning tasks.
\newblock In: \emph{Proceedings of the 26th ACM SIGKDD International Conference on Knowledge Discovery \& Data Mining}, KDD '20. New York, NY, USA: Association for Computing Machinery.
\newblock ISBN 9781450379984, p. 3561–3562.
\newblock \doi{10.1145/3394486.3406477}.
\newblock \urlprefix\url{https://doi.org/10.1145/3394486.3406477}.

\bibitem[{James~Hawdon and Räsänen(2017)}]{Hawdon2017}
James~Hawdon AO and Räsänen P (2017) Exposure to online hate in four nations: A cross-national consideration.
\newblock \emph{Deviant Behavior} 38(3): 254--266.
\newblock \doi{10.1080/01639625.2016.1196985}.
\newblock \urlprefix\url{https://doi.org/10.1080/01639625.2016.1196985}.

\bibitem[{Jha and Mamidi(2017)}]{jha2017does}
Jha A and Mamidi R (2017) When does a compliment become sexist? analysis and classification of ambivalent sexism using twitter data.
\newblock In: \emph{Proceedings of the second workshop on NLP and computational social science}. pp. 7--16.

\bibitem[{Kennedy et~al.(2022)Kennedy, Atari, Davani, Yeh, Omrani, Kim, Coombs, Havaldar, Portillo-Wightman, Gonzalez et~al.}]{kennedy2022introducing}
Kennedy B, Atari M, Davani AM, Yeh L, Omrani A, Kim Y, Coombs K, Havaldar S, Portillo-Wightman G, Gonzalez E et~al. (2022) Introducing the gab hate corpus: defining and applying hate-based rhetoric to social media posts at scale.
\newblock \emph{Language Resources and Evaluation} 56(1): 79--108.

\bibitem[{Kennedy et~al.(2020)Kennedy, Bacon, Sahn and von Vacano}]{kennedy2020constructing}
Kennedy CJ, Bacon G, Sahn A and von Vacano C (2020) Constructing interval variables via faceted rasch measurement and multitask deep learning: a hate speech application.
\newblock \emph{arXiv preprint arXiv:2009.10277} .

\bibitem[{{Landesanstalt für Medien NRW}(2023)}]{forsa2023}
{Landesanstalt für Medien NRW} (2023) {Forsa-Befragung zur Wahrnehmung von Hassrede}.
\newblock \url{https://www.medienanstalt-nrw.de/themen/hass/forsa-befragung-zur-wahrnehmung-von-hassrede.html}.
\newblock Accessed: 2023-11-15.

\bibitem[{Li and Caragea(2021)}]{li2021target}
Li Y and Caragea C (2021) Target-aware data augmentation for stance detection.
\newblock In: \emph{Proceedings of the 2021 Conference of the North American Chapter of the Association for Computational Linguistics: Human Language Technologies}. pp. 1850--1860.

\bibitem[{Liang et~al.(2022)Liang, Tadesse, Ho, Fei-Fei, Zaharia, Zhang and Zou}]{Liang2022}
Liang W, Tadesse GA, Ho D, Fei-Fei L, Zaharia M, Zhang C and Zou J (2022) {Advances, challenges and opportunities in creating data for trustworthy AI}.
\newblock \emph{Nat Machine Intelligence} 4: 669–677.
\newblock \doi{https://doi.org/10.1038/s42256-022-00516-1}.
\newblock \urlprefix\url{https://doi.org/10.1038/s42256-022-00516-1}.

\bibitem[{Lu et~al.(2020)Lu, Wu, Zhang, Zheng, Ren and Choo}]{lu2020cyberbullying}
Lu N, Wu G, Zhang Z, Zheng Y, Ren Y and Choo KKR (2020) Cyberbullying detection in social media text based on character-level convolutional neural network with shortcuts.
\newblock \emph{Concurrency and Computation: Practice and Experience} 32(23): e5627.

\bibitem[{Mandl et~al.(2019)Mandl, Modha, Majumder, Patel, Dave, Mandlia and Patel}]{mandl2019overview}
Mandl T, Modha S, Majumder P, Patel D, Dave M, Mandlia C and Patel A (2019) Overview of the hasoc track at fire 2019: Hate speech and offensive content identification in indo-european languages.
\newblock In: \emph{Proceedings of the 11th forum for information retrieval evaluation}. pp. 14--17.

\bibitem[{Maronikolakis et~al.(2022)Maronikolakis, Wisiorek, Nann, Jabbar, Udupa and Sch{\"u}tze}]{maronikolakis2022listening}
Maronikolakis A, Wisiorek A, Nann L, Jabbar H, Udupa S and Sch{\"u}tze H (2022) Listening to affected communities to define extreme speech: Dataset and experiments.
\newblock \emph{arXiv preprint arXiv:2203.11764} .

\bibitem[{Miceli et~al.(2021)Miceli, Yang, Naudts, Schuessler, Serbanescu and Hanna}]{miceli2021documenting}
Miceli M, Yang T, Naudts L, Schuessler M, Serbanescu D and Hanna A (2021) Documenting computer vision datasets: An invitation to reflexive data practices.
\newblock In: \emph{Proceedings of the 2021 ACM Conference on Fairness, Accountability, and Transparency}. pp. 161--172.

\bibitem[{Moy et~al.(2021)Moy, Raheem and Logeswaran}]{moy2021hate}
Moy TX, Raheem M and Logeswaran R (2021) Hate speech detection in english and non-english languages: A review of techniques and challenges.
\newblock \emph{Technology} .

\bibitem[{Nockleby(2000)}]{nockleby2000hate}
Nockleby JT (2000) Hate speech.
\newblock \emph{Encyclopedia of the American constitution} 3(2): 1277--1279.

\bibitem[{Ousidhoum et~al.(2019)Ousidhoum, Lin, Zhang, Song and Yeung}]{ousidhoum2019multilingual}
Ousidhoum N, Lin Z, Zhang H, Song Y and Yeung DY (2019) Multilingual and multi-aspect hate speech analysis.
\newblock \emph{arXiv preprint arXiv:1908.11049} .

\bibitem[{Page et~al.(2021)Page, McKenzie, Bossuyt, Boutron, Hoffmann, Mulrow, Shamseer, Tetzlaff, Akl, Brennan et~al.}]{page2021prisma}
Page MJ, McKenzie JE, Bossuyt PM, Boutron I, Hoffmann TC, Mulrow CD, Shamseer L, Tetzlaff JM, Akl EA, Brennan SE et~al. (2021) The prisma 2020 statement: an updated guideline for reporting systematic reviews.
\newblock \emph{International journal of surgery} 88: 105906.

\bibitem[{Pamungkas et~al.(2020)Pamungkas, Basile and Patti}]{pamungkas2020you}
Pamungkas EW, Basile V and Patti V (2020) Do you really want to hurt me? predicting abusive swearing in social media.
\newblock In: \emph{The 12th Language Resources and Evaluation Conference}. European Language Resources Association, pp. 6237--6246.

\bibitem[{Pamungkas et~al.(2023)Pamungkas, Basile and Patti}]{PamungkasBP23}
Pamungkas EW, Basile V and Patti V (2023) Towards multidomain and multilingual abusive language detection: a survey.
\newblock \emph{Pers. Ubiquitous Comput.} 27(1): 17--43.
\newblock \doi{10.1007/s00779-021-01609-1}.
\newblock \urlprefix\url{https://doi.org/10.1007/s00779-021-01609-1}.

\bibitem[{Paz et~al.(2020)Paz, Montero-Díaz and Moreno-Delgado}]{Paz2020}
Paz MA, Montero-Díaz J and Moreno-Delgado A (2020) Hate speech: A systematized review.
\newblock \emph{SAGE Open} 10(4): 2158244020973022.
\newblock \doi{10.1177/2158244020973022}.
\newblock \urlprefix\url{https://doi.org/10.1177/2158244020973022}.

\bibitem[{Pei and Jurgens(2023)}]{pei-jurgens-2023-annotator}
Pei J and Jurgens D (2023) When do annotator demographics matter? measuring the influence of annotator demographics with the {POPQUORN} dataset.
\newblock In: Prange J and Friedrich A (eds.) \emph{Proceedings of the 17th Linguistic Annotation Workshop (LAW-XVII)}. Toronto, Canada: Association for Computational Linguistics, pp. 252--265.
\newblock \doi{10.18653/v1/2023.law-1.25}.
\newblock \urlprefix\url{https://aclanthology.org/2023.law-1.25}.

\bibitem[{Poletto et~al.(2021)Poletto, Basile, Sanguinetti, Bosco and Patti}]{DBLP:journals/lre/PolettoBSBP21}
Poletto F, Basile V, Sanguinetti M, Bosco C and Patti V (2021) Resources and benchmark corpora for hate speech detection: a systematic review.
\newblock \emph{Lang. Resour. Evaluation} 55(2): 477--523.
\newblock \doi{10.1007/s10579-020-09502-8}.
\newblock \urlprefix\url{https://doi.org/10.1007/s10579-020-09502-8}.

\bibitem[{Qian et~al.(2019)Qian, Bethke, Liu, Belding and Wang}]{qian2019benchmark}
Qian J, Bethke A, Liu Y, Belding E and Wang WY (2019) A benchmark dataset for learning to intervene in online hate speech.
\newblock In: \emph{Proceedings of the 2019 Conference on Empirical Methods in Natural Language Processing and the 9th International Joint Conference on Natural Language Processing (EMNLP-IJCNLP)}. pp. 4755--4764.

\bibitem[{Risch et~al.(2021)Risch, Schmidt and Krestel}]{risch2021toxic}
Risch J, Schmidt P and Krestel R (2021) Data integration for toxic comment classification: Making more than 40 datasets easily accessible in one unified format.
\newblock In: \emph{Proceedings of the Workshop on Online Abuse and Harms (WOAH)}. Online: Association for Computational Linguistics, pp. 157--163.
\newblock \doi{10.18653/v1/2021.woah-1.17}.
\newblock \urlprefix\url{https://aclanthology.org/2021.woah-1.17}.

\bibitem[{R{\"o}ttger et~al.(2022)R{\"o}ttger, Nozza, Bianchi and Hovy}]{rottger-etal-2022-data}
R{\"o}ttger P, Nozza D, Bianchi F and Hovy D (2022) Data-efficient strategies for expanding hate speech detection into under-resourced languages.
\newblock In: \emph{Proceedings of the 2022 Conference on Empirical Methods in Natural Language Processing}. Abu Dhabi, United Arab Emirates: Association for Computational Linguistics, pp. 5674--5691.
\newblock \urlprefix\url{https://aclanthology.org/2022.emnlp-main.383}.

\bibitem[{Samory et~al.(2021)Samory, Sen, Kohne, Fl{\"o}ck and Wagner}]{samory2021call}
Samory M, Sen I, Kohne J, Fl{\"o}ck F and Wagner C (2021) Call me sexist, but…: Revisiting sexism detection using psychological scales and adversarial samples.
\newblock In: \emph{Proc. of the 15th International AAAI Conference on Web and Social Media (ICWSM)}. pp. 573--584.

\bibitem[{Sap et~al.(2022)Sap, Swayamdipta, Vianna, Zhou, Choi and Smith}]{sap-etal-2022-annotators}
Sap M, Swayamdipta S, Vianna L, Zhou X, Choi Y and Smith NA (2022) Annotators with attitudes: How annotator beliefs and identities bias toxic language detection.
\newblock In: Carpuat M, de~Marneffe MC and Meza~Ruiz IV (eds.) \emph{Proceedings of the 2022 Conference of the North American Chapter of the Association for Computational Linguistics: Human Language Technologies}. Seattle, United States: Association for Computational Linguistics, pp. 5884--5906.
\newblock \doi{10.18653/v1/2022.naacl-main.431}.
\newblock \urlprefix\url{https://aclanthology.org/2022.naacl-main.431}.

\bibitem[{Schmidt and Wiegand(2017)}]{schmidt-wiegand-2017-survey}
Schmidt A and Wiegand M (2017) A survey on hate speech detection using natural language processing.
\newblock In: \emph{Proceedings of the Fifth International Workshop on Natural Language Processing for Social Media}. Valencia, Spain: Association for Computational Linguistics, pp. 1--10.
\newblock \doi{10.18653/v1/W17-1101}.
\newblock \urlprefix\url{https://aclanthology.org/W17-1101}.

\bibitem[{Sen et~al.(2021)Sen, Fl{\"o}ck, Weller, Wei{\ss} and Wagner}]{sen2021total}
Sen I, Fl{\"o}ck F, Weller K, Wei{\ss} B and Wagner C (2021) A total error framework for digital traces of human behavior on online platforms.
\newblock \emph{Public Opinion Quarterly} 85(S1): 399--422.

\bibitem[{Siegel(2020)}]{siegel2020}
Siegel AA (2020) Online hate speech.
\newblock \emph{Social media and democracy: The state of the field, prospects for reform} : 56--88.

\bibitem[{Silva et~al.(2016)Silva, Mondal, Correa, Benevenuto and Weber}]{silva2016analyzing}
Silva L, Mondal M, Correa D, Benevenuto F and Weber I (2016) Analyzing the targets of hate in online social media.
\newblock In: \emph{Tenth international AAAI conference on web and social media}. pp. 687--690.

\bibitem[{Sodhi et~al.(2021)Sodhi, Pant and Mamidi}]{sodhi2021jibes}
Sodhi R, Pant K and Mamidi R (2021) Jibes \& delights: A dataset of targeted insults and compliments to tackle online abuse.
\newblock In: \emph{Proceedings of the 5th Workshop on Online Abuse and Harms (WOAH 2021)}. pp. 132--139.

\bibitem[{Swamy et~al.(2019)Swamy, Jamatia and Gamb{\"a}ck}]{swamy2019studying}
Swamy SD, Jamatia A and Gamb{\"a}ck B (2019) Studying generalisability across abusive language detection datasets.
\newblock In: \emph{Proceedings of the 23rd conference on computational natural language learning (CoNLL)}. pp. 940--950.

\bibitem[{Taradhita and Darma~Putra(2021)}]{taradhita2021hate}
Taradhita DAN and Darma~Putra I (2021) Hate speech classification in indonesian language tweets by using convolutional neural network.
\newblock \emph{Journal of ICT Research \& Applications} 14(3).

\bibitem[{Torregrosa et~al.(2021)Torregrosa, Bello-Orgaz, Martinez-Camara, Del~Ser and Camacho}]{torregrosa2021survey}
Torregrosa J, Bello-Orgaz G, Martinez-Camara E, Del~Ser J and Camacho D (2021) A survey on extremism analysis using natural language processing.
\newblock \emph{arXiv preprint arXiv:2104.04069} .

\bibitem[{Uyheng et~al.(2022)Uyheng, Moffitt and Carley}]{uyheng2022language}
Uyheng J, Moffitt J and Carley KM (2022) The language and targets of online trolling: A psycholinguistic approach for social cybersecurity.
\newblock \emph{Information Processing \& Management} 59(5): 103012.

\bibitem[{Vidgen and Derczynski(2020)}]{vidgen2020directions}
Vidgen B and Derczynski L (2020) Directions in abusive language training data, a systematic review: Garbage in, garbage out.
\newblock \emph{Plos one} 15(12): e0243300.

\bibitem[{Vidgen et~al.(2021)Vidgen, Nguyen, Margetts, Rossini, Tromble, Toutanova, Rumshisky, Zettlemoyer, Hakkani-Tur, Beltagy et~al.}]{vidgen2021introducing}
Vidgen B, Nguyen D, Margetts H, Rossini P, Tromble R, Toutanova K, Rumshisky A, Zettlemoyer L, Hakkani-Tur D, Beltagy I et~al. (2021) Introducing cad: the contextual abuse dataset.
\newblock In: \emph{Proceedings of the 2021 Conference of the North American Chapter of the Association for Computational Linguistics: Human Language Technologies}. Association for Computational Linguistics, pp. 2289--2303.

\bibitem[{Vigna et~al.(2017)Vigna, Cimino, Dell'Orletta, Petrocchi and Tesconi}]{del2017hate}
Vigna FD, Cimino A, Dell'Orletta F, Petrocchi M and Tesconi M (2017) Hate me, hate me not: Hate speech detection on facebook.
\newblock In: \emph{Proceedings of the First Italian Conference on Cybersecurity (ITASEC17), Venice, Italy, January 17-20, 2017}, \emph{{CEUR} Workshop Proceedings}, volume 1816. CEUR-WS.org, pp. 86--95.
\newblock \urlprefix\url{http://ceur-ws.org/Vol-1816/paper-09.pdf}.

\bibitem[{Vishwamitra et~al.(2020)Vishwamitra, Hu, Luo, Cheng, Costello and Yang}]{vishwamitra2020analyzing}
Vishwamitra N, Hu RR, Luo F, Cheng L, Costello M and Yang Y (2020) On analyzing covid-19-related hate speech using bert attention.
\newblock In: \emph{2020 19th IEEE International Conference on Machine Learning and Applications (ICMLA)}. IEEE, pp. 669--676.

\bibitem[{Waqas et~al.(2019)Waqas, Salminen, Jung, Almerekhi and Jansen}]{10.1371/journal.pone.0222194}
Waqas A, Salminen J, Jung Sg, Almerekhi H and Jansen BJ (2019) Mapping online hate: A scientometric analysis on research trends and hotspots in research on online hate.
\newblock \emph{PLOS ONE} 14(9): 1--21.
\newblock \doi{10.1371/journal.pone.0222194}.
\newblock \urlprefix\url{https://doi.org/10.1371/journal.pone.0222194}.

\bibitem[{Warner and Hirschberg(2012)}]{warner-hirschberg-2012-detecting}
Warner W and Hirschberg J (2012) Detecting hate speech on the world wide web.
\newblock In: \emph{Proceedings of the Second Workshop on Language in Social Media}. Montr{\'e}al, Canada: Association for Computational Linguistics, pp. 19--26.
\newblock \urlprefix\url{https://aclanthology.org/W12-2103}.

\bibitem[{Waseem et~al.(2017)Waseem, Davidson, Warmsley and Weber}]{waseem2017understanding}
Waseem Z, Davidson T, Warmsley D and Weber I (2017) Understanding abuse: A typology of abusive language detection subtasks.
\newblock \emph{arXiv preprint arXiv:1705.09899} .

\bibitem[{Waseem and Hovy(2016)}]{waseem2016hateful}
Waseem Z and Hovy D (2016) Hateful symbols or hateful people? predictive features for hate speech detection on twitter.
\newblock In: \emph{Proceedings of the NAACL student research workshop}. pp. 88--93.

\bibitem[{Wulczyn et~al.(2017)Wulczyn, Thain and Dixon}]{wulczyn2017ex}
Wulczyn E, Thain N and Dixon L (2017) Ex machina: Personal attacks seen at scale.
\newblock In: \emph{Proceedings of the 26th international conference on world wide web}. pp. 1391--1399.

\bibitem[{Zampieri et~al.(2019)Zampieri, Malmasi, Nakov, Rosenthal, Farra and Kumar}]{zampieri2019predicting}
Zampieri M, Malmasi S, Nakov P, Rosenthal S, Farra N and Kumar R (2019) Predicting the type and target of offensive posts in social media.
\newblock \emph{arXiv preprint arXiv:1902.09666} .

\bibitem[{Zhang et~al.(2018)Zhang, Robinson and Tepper}]{zhang2018detecting}
Zhang Z, Robinson D and Tepper J (2018) Detecting hate speech on twitter using a convolution-gru based deep neural network.
\newblock In: \emph{The Semantic Web: 15th International Conference, ESWC 2018, Heraklion, Crete, Greece, June 3--7, 2018, Proceedings 15}. Springer, pp. 745--760.

\end{thebibliography}
\end{document}